\pdfoutput=1

\documentclass[11pt]{article}

\usepackage[preprint]{acl}

\usepackage{times}
\usepackage{latexsym}
\usepackage{comment}
\usepackage[T1]{fontenc}

\usepackage[utf8]{inputenc}

\usepackage{microtype}

\usepackage{inconsolata}

\usepackage{graphicx}

\usepackage{booktabs}
\usepackage{multirow}
\usepackage{makecell}
\usepackage{amsmath}
\usepackage{tabularx}
\usepackage{marvosym}
\usepackage[ruled,vlined]{algorithm2e}
\usepackage{subcaption}

\hypersetup{
  pdftitle={BERAG: Bayesian Ensemble Retrieval-Augmented Generation for Knowledge-based Visual Question Answering},
  pdfauthor={Jinghong Chen, Jingbiao Mei, Guangyu Yang, Bill Byrne},
  pdfkeywords={retrieval-augmented generation, Bayesian ensemble, visual question answering}
}

\title{BERAG: Bayesian Ensemble Retrieval-Augmented Generation for Knowledge-based Visual Question Answering}

\author{Jinghong Chen, Jingbiao Mei, Guangyu Yang, Bill Byrne \\
        Department of Engineering \\ University of Cambridge \\ \texttt{\{jc2124, jm2245, gy266, wjb31\}@cam.ac.uk}}

\begin{document}
\maketitle
\begin{abstract}
Retrieval-Augmented Generation (RAG) is  used in  Knowledge-based Visual Question Answering (KB-VQA) to ground answers in external knowledge. This is commonly done by collecting multiple retrieved documents into a single context for answer generation.    This     can have drawbacks.  Concatenation   may not scale well to deep retrieval lists due to input context length constraints.  There may be  ``lost-in-the-middle'' effects, and the contributions of individual documents to answer generation can be unclear.
We propose Bayesian Ensemble Retrieval-Augmented Generation (BERAG) and Bayesian Ensemble Fine-Tuning (BEFT) as a RAG framework in which language models are conditioned on individual retrieved documents rather than a single combined context.  BERAG computes document posteriors  as ensemble weights that are updated token by token using Bayes' rule during answer  generation.
BERAG easily handles deep retrieval lists through its parallelizing document processing.  As a position-invariant ensemble method,  BERAG avoids ``lost-in-the-middle'' effects. BERAG document posteriors are strong indicators of individual document relevance.  We incorporate these posteriors into a Top-P inference pruning procedure, implemented in vLLM.  Relative to concatenative RAG, Top-P BERAG  offers reduced latency, faster inference, and strong VQA performance on  KB-VQA benchmarks, E-VQA and Infoseek, and long-context VQA tasks like SlideVQA and Multi-modal Needle-in-a-Haystack.
\end{abstract}

\section{Introduction}
\label{sec:introduction}

Retrieval Augmented Generation (RAG) is widely adopted to tackle Knowledge-based Visual Question Answering (KB-VQA). Grounding the answer generator, often a Vision Language Model (VLM), in retrieved documents that contain relevant knowledge is vital for KB-VQA question answering. Prior works often perform this grounding by concatenating all retrieved documents to form a single context, appending this context to the multi-modal query, and passing the long context to VLM for answer generation. While simple, this concatenative grounding approach has several limitations: (1) it is challenging to use deep retrieval lists of lengthy documents (e.g., Top-50 retrieved documents) for question answering due to maximal context length constraints and degraded performance when handling long context. A well-studied detrimental effect is known as ``lost-in-the-middle''~\citep{liu-etal-2024-lost}, where models neglect relevant information present in the middle of their context. Recent studies find that VLMs exhibit heightened sensitivity to positional arrangements compared to uni-modal settings~\citep{mm-lost-in-the-middle-whosinspotlight}. (2) It is difficult to determine how individual retrieved documents contribute to answer generation, in that the lack of document attribution poses challenges in verifying responses in practical deployment with human users.

To tackle these challenges, we revisit an \emph{ensemble} approach to  document grounding that was proposed in early RAG work~\citep{RAG-Lewis-NIPS2020}: \emph{``the top K documents are retrieved using the retriever, and the generator produces the output sequence probability for each document, which are then marginalized''}. We extend this approach by incorporating a Bayesian posterior probability over each document in response generation and proposing a training procedure that directly optimizes for this form of RAG generation. We term our inference procedure \textit{Bayesian Ensemble Retrieval-Augmented Generation (BERAG)}\footnote{Project page: \url{https://erichen0615.github.io/BERAG-project-page/}.} and our end-to-end training procedure \textit{Bayesian Ensemble Fine-Tuning (BEFT)}. We show this formalism eliminates ``lost-in-the-middle'' effects as all documents are treated as equal in the ensemble and also enables the use of much deeper retrieval lists under practical VRAM budgets. By modeling the document posterior, BERAG also explicitly quantifies document contribution, which we show can be useful for predicting whether retrieval is successful as well as for accelerating decoding.

Our experiments on E-VQA~\citep{EVQA-dataset} and Infoseek~\citep{Infoseek-dataset} show models trained with BEFT to generate answers following BERAG give strong performance compared to state-of-the-art systems. We  show that this gain derives from the ability to utilize the higher recall rate for deep retrieval whereas prior systems using concatenative grounding see degradation when given a long retrieval list. To our knowledge, these are the first systematic results on improving KB-VQA via utilizing very deep retrieval lists.
We further analyze the inference performance of BERAG systems and propose an acceleration technique based on the posterior distribution. With this, BERAG is substantially faster than concatenative RAG both in prefilling and in decoding. To summarize:

\begin{itemize}
    \item We propose BERAG and BEFT, an inference and end-to-end training framework for ensemble-based RAG (Sec.\ref{sec:method}).
    \item We show strong performance on E-VQA and Infoseek (Sec.\ref{sec:exp-E-VQA and Infoseek Visual Question Answering}, \ref{sec:exp-E-VQA and Infoseek Performance with respect to Context Length}) compared to recent systems. BERAG with our Top-P pruning is also faster than concatenative RAG in inference (Sec.~\ref{sec:BERAG Inference Latency}).
    \item We show BERAG resolves the ``lost-in-the-middle'' effect via position-invariant ensembling on KB-VQA tasks (Sec.\ref{sec:gt-doc-position-analysis}).
    \item As additional results, we show (1) BEFT learns to predict whether retrieval is successful; (2)
    BERAG performs well on long-context VQA tasks like SlideVQA and MMNeedle; and (3) the document posterior enables more fine-grained error attribution and thereby enhances interpretability. These are reported in the appendices.

\end{itemize}

These establish BERAG as a compelling alternative to the usual concatenative RAG for KB-VQA.

\section{Related Work}
\label{sec:related work}

\paragraph{Knowledge-based Visual Question Answering (KBVQA)} In KBVQA, the system is tasked to answer an image-related question based on knowledge in retrieved documents. E-VQA~\cite{EVQA-dataset} and Infoseek~\citep{Infoseek-dataset} are widely used Wikipedia-based KBVQA benchmarks, covering knowledge about animals and plants, as well as facts on sites and entities.
Recent progress has been driven primarily by stronger retrieval:
PreFLMR~\citep{FLMR} shows that fine-grained late interaction substantially improves document retrieval for KBVQA, EchoSight~\citep{EchoSight} introduces a dedicated mult-modal reranker to improve retrieval, and ReflectiVA~\citep{ReflectiVA} adds reflective tokens to the model for deciding whether to retrieve and evaluating passage relevance.

\paragraph{Joint optimisation of retrieval and answer generation in RAG} Previous works focus on joint retriever-answer generator training~\citep{ReAuSE, RetGen, RAVQA}. Most relevant to our work is RetGen~\citep{RetGen}, which  introduces Actor-Critic style joint training objective to include retriever score by considering the marginalization in Eq.\ref{eq:BERAG-probability}.
In contrast to their focus on retriever-generator joint optimization, we train a single probabilistic answer generation system.
This formalism yields supervised fine-tuning objectives with token-level supervision signals for document weighting, whereas RetGen provides a sequence-level correction signal for the retriever. We discuss other related work in Appendix \ref{app:extended_related_work}.

\section{Method}
\label{sec:method}
\subsection{Bayesian Ensemble Retrieval Augmented Generation (BERAG)}

The Retrieval Augmented Generation objective, $\arg\max_{\bf y} P({\bf y} | {\bf x}, Z) $, is to generate response $\bf y$ given a query $\bf x$ and a collection of retrieved documents $Z=[z_1,\ldots,z_K]$. The query $\mathbf{x}$ can be multi-modal. Let $\Theta$ denote the set of all model parameters.
We now describe how Standard RAG and our method, Bayesian Ensemble RAG, compute the $Z$-conditioned likelihood $P({\bf y} | {\bf x}, Z)$.

\paragraph{Concatenative RAG}  Conditioning on $Z$ is done by concatenating the documents $z_1,\ldots,z_K$, often in the same order as they are retrieved, to the query $\bf x$ to form the input context. Let $\circ$ denote concatenation. We have

\begin{align}
P({\bf y} | {\bf x}, Z)  \simeq   P_\Theta({\bf y} | {\bf x} \circ z_1 \circ \ldots \circ z_K)
\end{align}
\noindent where for $\mathbf{y} = y_1, \ldots y_n$ , probability is computed as
$
 P_\Theta({\bf y} | {\bf x} \circ z_1 \circ \ldots \circ z_K)  = \prod_{i=1}^n  P_\Theta( y_i | y_{<i}, {\bf x} \circ z_1 \circ \ldots \circ z_K)
$.

\paragraph{Bayesian Ensemble RAG} Conditioning on $Z$ is done by marginalizing over all singleton documents
\begin{align}
    &P({\bf y} | {\bf x}, Z) = \prod_{j=1}^n P(y_j | y_{<j}, {\bf x}, Z) \notag \\
    &\simeq \prod_{j=1}^n
\sum_{k=1}^K P(y_j,z_k|y_{<j},\mathbf{x},Z) \notag \\
&= \prod_{j=1}^n\sum_{k=1}^K P_\Theta(y_j|y_{<j},\mathbf{x},z_k)P_\Theta(z_k|y_{<j},\mathbf{x}, Z)
\label{eq:BERAG-probability}
\end{align}
\noindent where the \textit{document posterior} $P(z_k|y_{<j},\mathbf{x},Z)$ is computed via Bayes' rule~\citep{RetGen, danielle-domain-adaptive-nmt, allauzen2011bayesian}:
{\small
\begin{align}
&P(z_k|y_{<j},\mathbf{x}, Z) \notag
\\&=\frac{P_\Theta(y_{<j}|z_k,\mathbf{x})P_\Theta(z_k|\mathbf{x}, Z)}{\sum_{k'=1}^K P_\Theta(y_{<j} | z_{k'},\mathbf{x})P_\Theta(z_{k'}|\mathbf{x},Z)}  \label{eq:berag document posterior} \\
&=\frac{P_\Theta(y_{j-1}|y_{<j-1}, z_k,\mathbf{x}) P(z_k|y_{<j-1},\mathbf{x}, Z)}{\sum_{k'=1}^K P_\Theta(y_{j-1}|y_{<j-1}, z_{k'}, \mathbf{x}) P(z_{k'}|y_{<j-1},\mathbf{x}, Z)}  \label{eq:berag posterior recursive form}
\end{align}
}
Eq.~\ref{eq:berag document posterior} is obtained via applying Bayes' rule to the posterior. The recursive Eq.~\ref{eq:berag posterior recursive form} (derived in Appendix \ref{app:derivation}) is useful in inference for avoiding repeated computation of the whole sequence likelihood $P_\Theta(y_{<j} |z_k, \mathbf{x})$.

The \textit{document prior} $P(z_k|\mathbf{x}, Z)$, that is, the base case of the recursion, is computed as follows: let $\mathbf{e}_\Theta(\mathbf{x}, z_k)$ denote the last-layer embedding for the last input token (e.g., <EOS>) from the language model $\Theta$ given query $\mathbf{x}$ and document $z_k$. We use this embedding as input to a trainable two-layer MLP $s(\cdot)$ to compute the document prior
\begin{equation}
    P_\Theta(z_k|\mathbf{x}, Z)=\frac{\exp s(\mathbf{e}_\Theta(\mathbf{x}, z_k))}{\sum_{k'=1}^K \exp s(\mathbf{e}_\Theta(\mathbf{x}, z_{k'}))}
    \label{eq:berag document prior}
\end{equation}

We term the above inference procedure \textbf{Bayesian Ensemble Retrieval-Augmented Generation (BERAG)} as it performs ensemble inference over individual documents during RAG with Bayesian weight update. The procedure is illustrated in Figure \ref{fig:berag-system-diagram} in the appendix. We note that BERAG differs from previous works in that (1) document posterior is computed via Bayes' rule; and (2) document prior is computed via an MLP layer rather than normalizing document scores assigned by a  retriever. Detailed comparison with related methods is discussed in Sec.\ref{sec:related work}.

\subsection{Bayesian Ensemble Fine-Tuning (BEFT)}

We now show how to fine-tune a BERAG system end-to-end.
A RAG training set of $N$ items is defined as $\mathcal{D}=\left\{\big(\mathbf{x}^i,\mathbf{y}^i,Z^i\big):Z^i=\big[z^{i}_{1},\ldots,z^{i}_{K}\big]\right\}_{i=1}^{N}$.
We denote ground-truth document relevance labels as $r^{i}_{k} \in \{0,1\}$ when available. $r^{i}_{k}$ indicates whether the $k^{th}$ document $z^{i}_{k}\in Z^i$ is relevant to generating the answer ${\bf y}^i$ in response to the query ${\bf x}^i$.  BEFT minimizes negative log-likelihood loss based on Eq.\ref{eq:BERAG-probability} over $\mathcal{D}$.
\begin{align}
    &\mathcal{L}_{BEFT}(\Theta)=-\sum_{i=1}^N \log P(\mathbf{y}^i|\mathbf{x}^i,Z^i) \notag \\
    &=-\sum_{i=1}^N \sum_{j=1}^{n_i}\log\bigg(\sum_{k=1}^K P(y_j^i,z_k^i|y_{<j}^i,\mathbf{x}^i,Z^i)\bigg) \notag \\
&= -\sum_{i=1}^N \sum_{j=1}^{n_i}\log\bigg(
\sum_{k=1}^K
\underbrace{P_\Theta(y_j^i \mid \mathbf{x}^i, z_k^i, y_{<j}^i)}_{\text{Next-Token Likelihood}}
\cdot \notag \\
& \quad\quad
\underbrace{
\frac{
P_\Theta(y_{<j}^i \mid \mathbf{x}^i, z_k^i)\,P_\Theta(z_k^i \mid \mathbf{x}^i, Z^i)
}{
\sum_{k'=1}^K
P_\Theta(y_{<j}^i \mid \mathbf{x}^i, z_{k'}^i)\,P_\Theta(z_{k'}^i \mid \mathbf{x}^i, Z^i)
}
}_{\text{Document Posterior $P(z_k^i|y_{<j}^i,\mathbf{x}^i,Z^i)$ (Eq.\ref{eq:berag document posterior})}}
\bigg)
    \label{eq:BEFT-loss}
\end{align}
\noindent where $n_i$ denotes the length of the target answer $\mathbf{y}^i$ and $P(z_{k^{'}}^{i}|\mathbf{x}^i,Z^i)$ is initialised according to Eq.\ref{eq:berag document prior}. The BEFT loss in Eq.~\ref{eq:BEFT-loss} can be viewed as a weighted maximum likelihood objective, where the document prior and posterior weights control the contribution from each document. BEFT does not require the ground-truth document relevance labels $r_k^i$ (their use is discussed in appendix~\ref{sec:SLDP}).

\paragraph{BEFT Gradient} We find empirically that BEFT learns a strong prior distribution. This can be explained by the gradient of the BEFT loss. It can be shown that the next-token likelihood modulates how BEFT maximizes the document posterior.  If token $y_j$ is more likely under document $z_{1}$ than under document $z_{2}$, BEFT allocates a larger gradient magnitude toward maximizing the posterior
$P_\Theta(z_{1} \mid y_{<j}, \mathbf{x}, Z)$ than toward maximizing $P_\Theta(z_{2} \mid y_{<j}, \mathbf{x}, Z)$.

Because the document posterior is related to the document prior through Bayes' rule, it can further be shown that the gradient with respect to the prior probabilities is proportional to the posterior probabilities. This suggests that BEFT can be interpreted as a form of self-supervision that learns evidence selection directly from question-answering training. The supervision signal is provided by the model's own token-level log-likelihood. Full derivation is provided in Appendix~\ref{app:BEFT Update gradients}.

\section{Experiments on KB-VQA}
\label{sec:experiments}

\subsection{Experimental Setup}
\label{sec:experimental setup}

\paragraph{Datasets} We experiment on Encyclopedic-VQA (E-VQA) ~\cite{EVQA-dataset} and  Infoseek~\cite{Infoseek-dataset}. For training, we use the train-test splits and passage sets as compiled in the Multi-Modal Knowledge Retrieval (M2KR) Benchmark~\citep{PreFLMR}. For inference, we report the original and the M2KR versions of E-VQA to compare against prior works. The M2KR version has 150K passages whereas the original version has 2M passages. For Infoseek, we report on the M2KR test split which contains 4.7K examples derived from the original full validation set, and additionally report BEFT results on the full validation set of 71,335 examples.

\paragraph{Evaluation} We use Recall@K to evaluate retrieval based on the official ground-truth document labels in the E-VQA and Infoseek datasets, \textit{not} Pseudo-Recall. For VQA performance, we use the official VQA metrics, BERT Match (BEM) and VQA score, averaged over the test set for evaluation on E-VQA and Infoseek, respectively.

\paragraph{Models} We use the Qwen2-VL-Instruct~\citep{qwen2vl-technical-report} and Qwen3-VL-Instruct~\citep{qwen3vltechnicalreport-2025} families of VLMs for answer generation. We also apply BEFT to Llava-Llama-3~\citep{llava-llama-model} on E-VQA. For retrieval, we use PreFLMR-L (442M) and EVA-CLIP-8B following prior works.

\paragraph{BEFT Training} We provide $K=2$ documents for BEFT training on E-VQA, Infoseek, and MMNeedle, where one document is the correct grounding. We set $K=4$ for SlideVQA and include the prior loss (Eq.~\ref{eq:beft_prior_loss}), so that all gold documents can be included.

\paragraph{Fine-tuned Baselines} We establish Supervised Fine-tuning (SFT) and Direct Preference Optimization (DPO) baselines for comparison with BEFT. Following prior work, we conduct SFT with the Top-5 documents from PreFLMR-L retrieval in context~\citep{RAVQA}. The Top-5 might not contain the ground-truth document. For DPO, we generate 8 samples per question using the SFT model conditioned on the Top-5 documents, and choose a correct response as preferred and an incorrect response as dis-preferred to form pairwise preference data following \citet{dpombr-naacl}. DPO training is performed on the SFT model. We perform a grid-search to identify the optimal $\beta$ value for DPO.
Full details are provided in Appendix \ref{app:experimental_setup_details}.

\subsection{E-VQA and Infoseek Visual Question Answering}
\label{sec:exp-E-VQA and Infoseek Visual Question Answering}

\begin{table*}[!ht]
\centering
\small
\setlength{\tabcolsep}{5pt}
\renewcommand{\arraystretch}{0.95}
\setlength{\aboverulesep}{0.3ex}
\setlength{\belowrulesep}{0.45ex}
\begin{tabular}{@{}lllcr@{}}
\toprule
\# & \textbf{System} & \textbf{Retriever (R@5)}
& \textbf{Answer Generator} & \textbf{Best score} \\
\midrule
\multicolumn{5}{c}{\textbf{E-VQA~[M2KR, 150K passages]}} \\
\midrule
1 & UniIR~\citep{UniIR}
  & UniIR (22.0) & T5-Large & 25.7 \\
2 & ReT~\citep{RETpaper}
  & ReT (46.3) & LLaMA-3.1-8B & 43.7 \\
3 & RAVQA-v2~\citep{PreFLMR}
  & PreFLMR-L (60.6) & T5-Large & 54.5 \\
4 & MuKA~\citep{deng-etal-2025-muka}
  & MuKA (64.9) & VILA-13B & 63.1 \\
5 & MM-RewardRAG~\citep{mmreward-rag-emnlp-finding}
  & PreFLMR-G (62.4) & Qwen2-VL-7B-Instruct & 63.3 \\
\midrule
6 & GPT-4o-mini
  & PreFLMR-L (60.6) & GPT-4o-mini & 63.8 \\
7 & SFT
  & PreFLMR-L (60.6) & Llava-Llama-3-8B & 57.8 \\
8 & SFT
  & PreFLMR-L (60.6) & Qwen2-VL-7B-Instruct & 61.5 \\
9 & SFT
  & PreFLMR-L (60.6) & Qwen3-VL-8B-Instruct & 59.4 \\
10 & DPO (ours)
  & PreFLMR-L (60.6) & Qwen2-VL-7B-Instruct & 64.1 \\
11 & BEFT (ours)
   & PreFLMR-L (60.6) & Llava-Llama-3-8B & 63.7 \\
12 & BEFT (ours)
   & PreFLMR-L (60.6) & Qwen2-VL-7B-Instruct & 70.3 \\
13 & BEFT (ours)
   & PreFLMR-L (60.6) & Qwen3-VL-8B-Instruct & \textbf{71.8} \\
   \midrule
\multicolumn{5}{c}{\textbf{E-VQA~[Original, 2M Passages]}} \\
\midrule
14 & ReflectiVA~\citep{ReflectiVA}
   & EVA-CLIP-8B & LLaMA-3.1-8B & 35.5 \\
15 & VLM-PRF~\citep{hong2025knowledge}
   & EVA-CLIP-8B & InternVL3-8B & 40.1 \\
16 & CC-VQA~\citep{hong2026ccvqa}
   & EVA-CLIP-8B & Qwen2.5-VL-7B-Instruct & 41.4 \\
17 & REAL~\citep{ye2026real}
   & EVA-CLIP-8B & Qwen3-VL-8B-Instruct & 45.5 \\
18 & BEFT (ours)
   & EVA-CLIP-8B & Qwen3-VL-8B-Instruct & \textbf{46.1} \\
   \midrule
\multicolumn{5}{c}{\textbf{Infoseek~[100K Passages]}} \\
\midrule
19 & UniIR~\citep{UniIR}
   & UniIR & T5-Large & 26.4 \\
20 & RAVQA-v2~\citep{PreFLMR}
   & PreFLMR-L & T5-Large & 33.5 \\
21 & ReflectiVA~\citep{ReflectiVA}
   & EVA-CLIP-8B & LLaMA-3.1-8B & 40.1 \\
22 & MuKA~\citep{deng-etal-2025-muka}
   & MuKA & VILA-13B & 41.0 \\
23 & EchoSight~\citep{EchoSight}
   & EVA-CLIP-8B & LLaMA-3-8B & 41.8 \\
24 & MM-RewardRAG~\citep{mmreward-rag-emnlp-finding}
   & PreFLMR-L & Qwen2-VL-7B-Instruct & 44.4 \\
25 & MMSearch-R1~\citep{wu2025mmsearchr1}
   & BGE + EVA-CLIP & Qwen2.5-VL-7B-Instruct & 39.7 \\
26 & DeepEyesV2~\citep{hong2026deepeyesv2}
   & BGE + EVA-CLIP & Qwen2.5-VL-7B-Instruct & 41.3 \\ \midrule
27 & CC-VQA~\citep{hong2026ccvqa}
   & EVA-CLIP-8B & Qwen2.5-VL-7B-Instruct & 45.1$^{\dagger}$ \\
28 & REAL~\citep{ye2026real}
   & EVA-CLIP-8B & Qwen3-VL-8B-Instruct & 44.1$^{\dagger}$ \\
29 & MaS-VQA~\citep{mao2026masvqa}
   & EVA-CLIP-8B & Qwen3-VL-8B-Instruct & 43.8$^{\dagger}$ \\
30 & BEFT (ours)
   & EVA-CLIP-8B & Qwen2-VL-7B-Instruct & 43.5$^{\dagger}$ \\
31 & BEFT (ours)
   & EVA-CLIP-8B & Qwen3-VL-8B-Instruct & \textbf{45.4}$^{\dagger}$ \\
\midrule
32 & SFT
   & EVA-CLIP-8B (61.9) & Qwen2-VL-7B-Instruct & 40.8 \\
33 & SFT
   & EVA-CLIP-8B (61.9) & Qwen3-VL-8B-Instruct & 36.8 \\
34 & DPO (ours)
   & EVA-CLIP-8B (61.9) & Qwen2-VL-7B-Instruct & 39.7 \\
35 & BEFT (ours)
   & PreFLMR-L (36.6) & Qwen2-VL-7B-Instruct & 42.8 \\
36 & BEFT (ours)
   & EVA-CLIP-8B (61.9) & Qwen2-VL-7B-Instruct & \textbf{45.1} \\
37 & BEFT (ours)
   & EVA-CLIP-8B (61.9) & Qwen3-VL-8B-Instruct & 44.7 \\
\bottomrule
\end{tabular}
\caption{VQA performance on E-VQA and Infoseek datasets.
For comparison to the literature,  we report E-VQA performance on both the M2KR subset and the full original evaluation set.
We identify the retriever used by each system and note its Recall@5 (R@5) when available.  Best performance is not necessarily achieved at
K = 5. $^{\dagger}$ denotes scores evaluated on the full Infoseek validation set, which is a superset of the M2KR test split, with the same 100K passage set. Full-validation BEFT results use $K=5$. We emphasize BEFT models generate answers without reasoning prefix.}
\label{tab:best_vqa_performance_against_baselines}
\end{table*}

Table \ref{tab:best_vqa_performance_against_baselines} compares our SFT, DPO, and BEFT models against recent works on E-VQA and Infoseek.

\paragraph{BEFT Performance} On the M2KR E-VQA set, where passages are chunked into sections and which we use for training, BEFT surpasses strong baseline systems and prior works by large margins. Relative to their SFT baselines, we see $>5.0$ gains in VQA score by applying BEFT to Llava-Llama, Qwen2-VL, and Qwen3-VL. The Qwen3-VL-based BEFT model trained on sections generalizes well to the original E-VQA, where the whole article is provided and where retrieval is more difficult given the large passage set. We note that BERAG produces answers  without the need of reasoning prefix, as in standard RAG. This contrasts with CC-VQA~\citep{hong2026ccvqa} and REAL~\citep{ye2026real} which makes use of reasoning guidance. This makes BERAG simple and fast in inference (Sec.~\ref{sec:BERAG Inference Latency}). On Infoseek, BEFT achieves 45.1 on the M2KR test split and 45.4 on the full validation set, comparable to state-of-the-art systems.

\begin{table*}[ht!]
\centering
\small
\setlength{\tabcolsep}{6pt}
\begin{tabular}{llccccccccc}
\toprule
 &  & \multicolumn{9}{c}{\textit{E-VQA}} \\
\midrule
\# &  & \textbf{K=1} & \textbf{K=2} & \textbf{K=3} & \textbf{K=5} & \textbf{K=10} & \textbf{K=15} & \textbf{K=20} & \textbf{K=30} & \textbf{K=50} \\ \midrule
 & \textbf{Recall@K} & 39.5\% & 46.9\% & 56.6\% & 60.6\% & 68.0\% & 72.3\% & 75.0\% & 81.3\% & 84.4\% \\
\midrule
1 & GPT-4o-mini & \textbf{54.2} & \textbf{59.4} & \textbf{60.4} & 61.6 & 63.0 & 63.6 & 63.8 & 60.2 & 61.3 \\ \midrule
2 & Qwen2-VL-7B-Instruct & \underline{51.6} & 55.9 & 54.7 & 55.1 & 53.9 & 56.8 & 56.0 & 54.2 & \textit{OOL} \\
3 & \quad SFT & 51.5 & 55.7 & 57.1 & 58.6 & 58.4 & 61.5 & 60.5 & 59.1 & \textit{OOL} \\
4 & \quad BEFT (ours) & 50.4 & 56.0 & 59.9 & \underline{63.4} & \underline{67.1} & \underline{68.7} & \underline{69.4} & \underline{69.8} & \underline{70.3} \\
\midrule
5 & Qwen3-VL-8B-Instruct & 50.4 & 56.1 & 55.6 & 57.9 & 60.1 & 60.1 & 60.0 & 60.0 & 59.2 \\
6 & \quad SFT & \underline{51.2} & 56.5 & 56.3 & 59.4 & 58.2 & 57.8 & 59.3 & 58.2 & 55.5 \\
7 & \quad BEFT (ours) & 50.6 & \underline{57.2} & \underline{60.0} & \textbf{63.4} & \textbf{67.7} & \textbf{69.1} & \textbf{70.0} & \textbf{71.1} & \textbf{71.8} \\
\midrule
\multicolumn{2}{l}{} & \multicolumn{9}{c}{\textit{Infoseek}} \\
\midrule
\# &  & \textbf{K=1} & \textbf{K=2} & \textbf{K=3} & \textbf{K=5} & \textbf{K=10} & \textbf{K=15} & \textbf{K=20} & \textbf{K=30} & \textbf{K=50} \\ \midrule
 & \textbf{Recall@K} & 38.4\% & 55.6\% & 57.7\% & 61.9\% & 69.7\% & 72.7\% & 75.1\% & 77.8\% & 79.7\% \\
\midrule
 & Qwen2-VL-7B-Instruct & \multicolumn{9}{c}{} \\
8 & \quad SFT & \textbf{39.6} & 40.5 & 40.8 & 40.8 &
\multicolumn{5}{c}{\textit{----------------------- OOL -----------------------}\hspace{0.4em}\leaders\hrule height 0.4pt\hfill} \\
9 & \quad BEFT & \underline{38.6} & \textbf{42.8} & \textbf{43.9} & \textbf{45.1} & \textbf{44.9} & \underline{43.7} & \underline{43.1} & \underline{42.3} & \underline{41.6} \\
\midrule
 & Qwen3-VL-8B-Instruct & \multicolumn{9}{c}{} \\
10 & \quad SFT & 36.3 & 37.0 & 37.2 & 36.4 & 36.5 & 37.0 & 36.8 & 36.8 & 37.5 \\
11 & \quad BEFT & 38.4 & \underline{41.2} & \underline{43.4} & \underline{44.3} & \underline{44.7} & \textbf{44.4} & \textbf{43.6} & \textbf{43.2} & \textbf{42.2} \\
\bottomrule
\end{tabular}
\caption{VQA performance of BEFT, SFT, and base models for different retrieval Top-$K$, where documents are retrieved by PreFLMR-L.
E-VQA is evaluated with Bert Exact Match (BEM) and Infoseek with Exact Match (EM) following the official evaluation script.
Qwen2-VL and Qwen3-VL without fine-tuning, as well as GPT-4o-mini, give low scores on Infoseek under Exact Match even when prompted to generate short answers and so we do not report them.
BEFT models consistently generate better answers with longer context, making use of the higher recall rate at large $K$.
The 32K context window for Qwen2-VL-7B-Instruct is exceeded at $K>30$ on E-VQA and $K>15$ on Infoseek under regular decoding.
\textit{OOL} denotes Out Of Context Length. \textbf{Bold} denotes best performance at the operating Top-$K$.}
\label{tab:vqa_results}
\end{table*}

\begin{figure*}[!ht]
    \centering
    \begin{subfigure}[t]{0.49\textwidth}
        \centering
        \includegraphics[width=\linewidth]{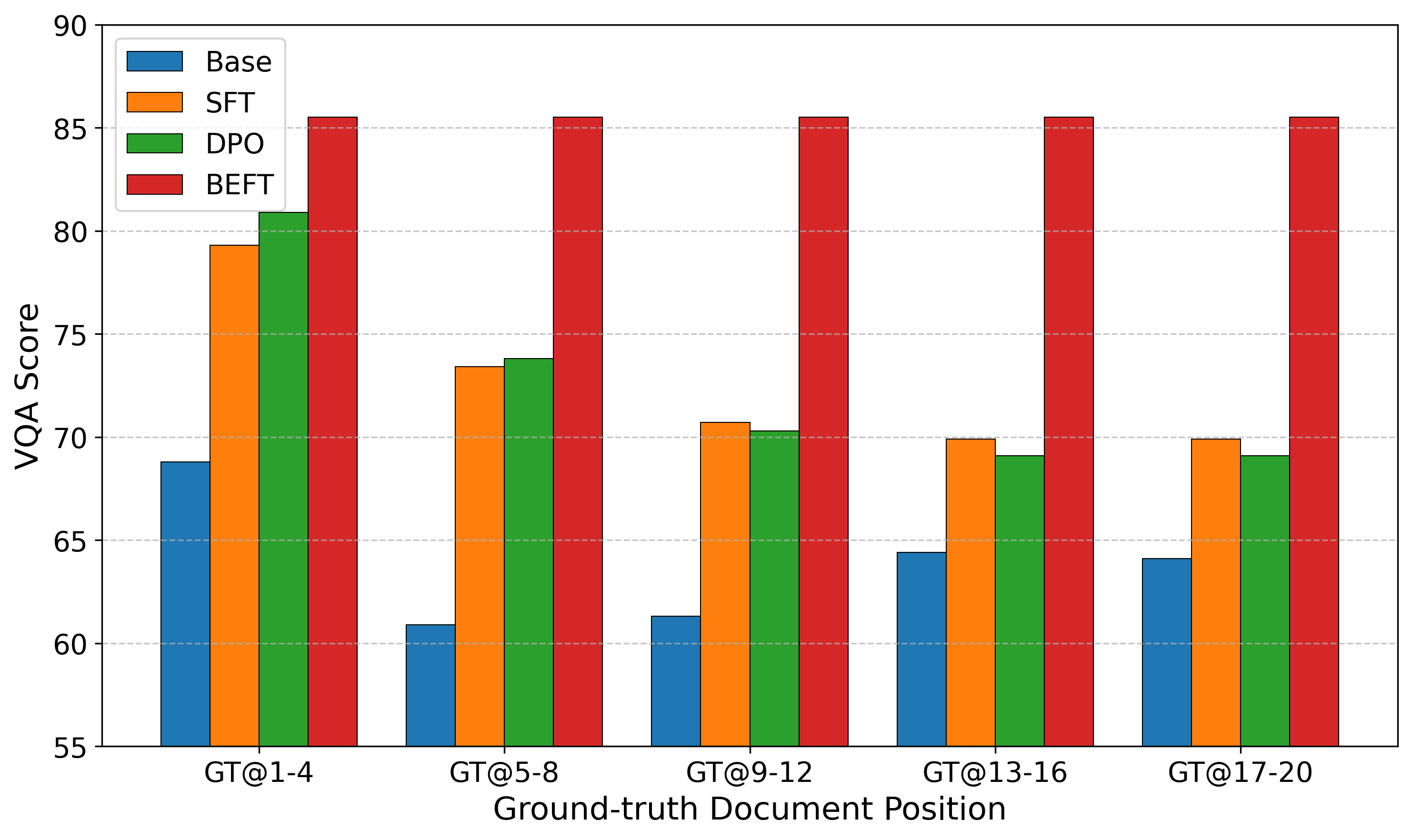}
        \caption{Qwen2-VL-7B-Instruct.}
        \label{fig:GTDocPosition-Qwen2}
    \end{subfigure}
    \hfill
    \begin{subfigure}[t]{0.49\textwidth}
        \centering
        \includegraphics[width=\linewidth]{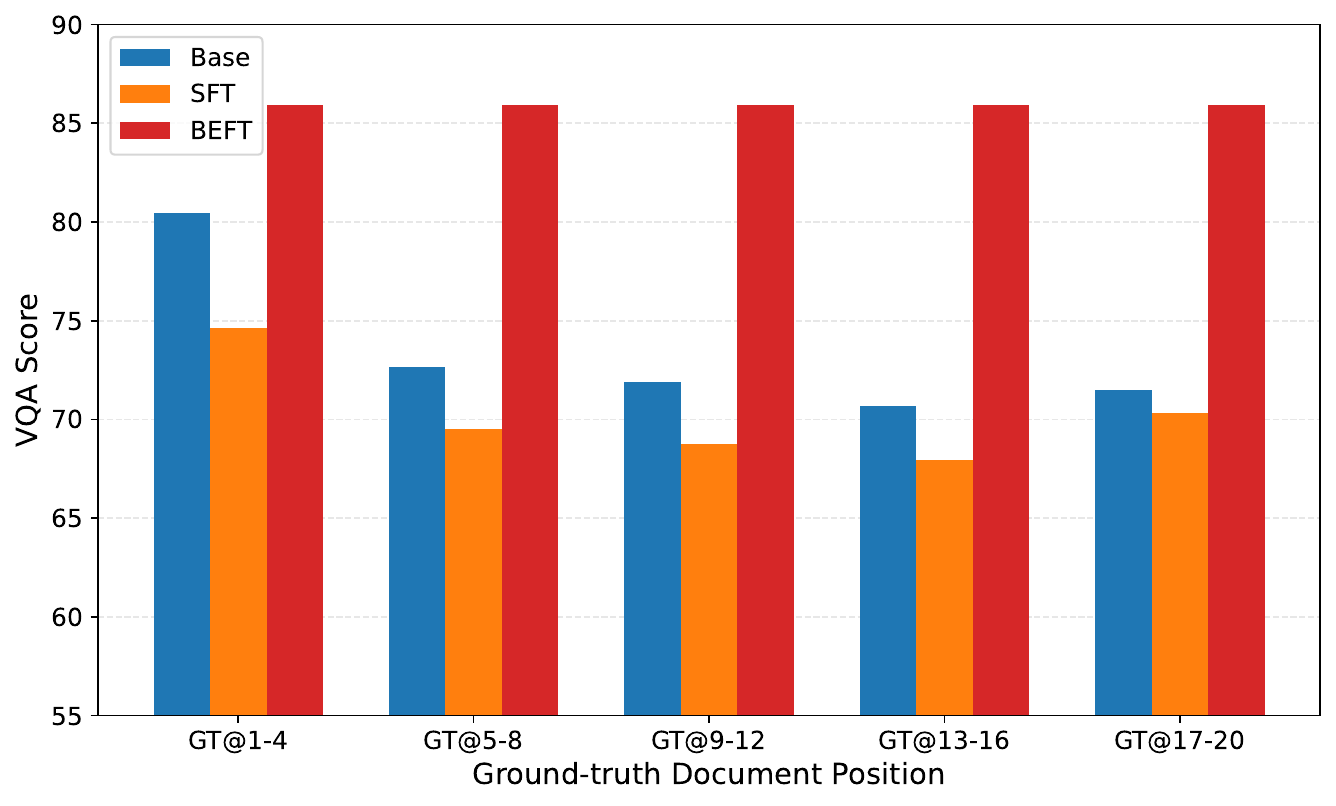}
        \caption{Qwen3-VL-8B-Instruct.}
        \label{fig:GTDocPosition-Qwen3}
    \end{subfigure}

    \caption{VQA scores at different ground-truth document (GT doc) positions in the retrieval list. Both panels are evaluated on the same set of 256 randomly sampled E-VQA questions for which the GT doc occurs in the Top-20 list retrieved by PreFLMR-L. GT@1--4 indicates that the GT doc is moved into the Top-4 by swapping documents when necessary, and vice versa. Both Qwen2-VL and Qwen3-VL see severe degradations (>-10) when the gold documents are not placed in the Top-4, whereas BEFT models (red) are unaffected by gold document positioning.}
    \label{fig:GTDocPosition-vs-VQAScore}
\end{figure*}

\subsection{E-VQA and Infoseek Performance with respect to Context Length}
\label{sec:exp-E-VQA and Infoseek Performance with respect to Context Length}

Table \ref{tab:vqa_results} reports VQA performance of the base model, our fine-tuning baselines and BEFT models on E-VQA and Infoseek for Qwen2-VL-7B and Qwen3-VL-8B. We report the retriever's Recall@$K$ and system performance for $K$ in the range of $1$ to $50$. This allows us to understand how these systems scale with growing input context.

\paragraph{Best Performance is achieved at Moderate to Large K.}
In the small K regime, the recall rates are low (e.g., <60\%), meaning that in many instances, the provided documents are not sufficient for question answering. This challenges the model to rely on its own parametric knowledge. We see this behavior in E-VQA, K=1,2,3,5, where VQA score can be higher than the recall rate. In the large K regime, the recall rates increase substantially to >70\%, and the challenge now is to identify the relevant document among many confounders and generate answers accordingly. Overall, the optimal performance is achieved at the moderate range ($K\in [5,20]$).

\paragraph{Pretrained Qwen3-VL models handle deep retrieval lists better than Qwen2-VL models.} For E-VQA, Qwen2-VL (row 2) struggles to utilize deep retrieval lists, restricted by its limited context length and the degradation at long context. On Infoseek, Qwen2-VL is unable to handle depths $K\geq 10$; in comparison, Qwen3-VL (row 5) has a much longer maximal context length and maintains performance at long contexts, although VQA score also peaks at $K=15$.

\paragraph{Qwen3-VL and Qwen2-VL perform similarly after SFT.}
Although the pretrained Qwen3-VL model achieves much higher score on E-VQA relative to the Qwen2-VL pretrained model, they give similar performance after SFT.  Allowing for variations in retrieval depth, Qwen2-VL achieves higher performance than Qwen3-VL on both E-VQA and Infoseek after SFT.
This may well be attributed to Qwen3-VL's extensive reinforcement learning-based post training, which could make SFT less effective.

\paragraph{BEFT enhances performance substantially for both the Qwen2-VL and the Qwen3-VL models.
}
When given multiple documents, the BEFT models outperform baselines at all retrieval depths.  We report BEFT  at $K=1$, although we note that BEFT is not intended for single documents.  We find that BEFT makes good use of Qwen2-VL at retrieval depths that the model could otherwise not accommodate. Although Qwen3-VL is the more recent model, on Infoseek we find that BEFT with Qwen2-VL gives the slightly better best performance ($+0.5\%$). However, on E-VQA, BEFT with Qwen3-VL is clearly the stronger system.

\subsection{Effect of Ground-truth Document Position on RAG and BERAG}
\label{sec:gt-doc-position-analysis}

When generating with long context under standard RAG, Large Language Models are known to be sensitive to where the relevant information is placed. \citet{liu-etal-2024-lost} find that ``\textit{Model performance is highest when relevant information occurs at the beginning or end of its input context}''. In Figure \ref{fig:GTDocPosition-vs-VQAScore}, we evaluate systems based on Qwen2-VL and Qwen3-VL to investigate whether recent models are still sensitive to gold document positioning. To do this, we form 5 subsets where the gold document is placed at different positions (e.g., GT@5-8 indicates the gold document index is between 5 and 8, inclusive) from the same set of 256 E-VQA questions. Ideally, systems should give identical scores on these subsets as they only vary in document positioning.

The Qwen2-VL base model is clearly ``Lost-in-the-Middle'', with low performance when the GT doc is placed between 5th-8th and 9th-12th in the Top-20 list. SFT and DPO do not correct for this behavior. In comparison, BEFT gives the same and highest VQA score among the four systems, as expected given the ensemble formalism in Eq.~\ref{eq:BERAG-probability} is unaware of document position. The Qwen3-VL base model gives stronger baseline performance compared to Qwen2-VL, but still strongly favors having the gold document in the Top-4. Again, the BEFT model gives identical scores not affected by gold document placement.

\subsection{Efficient BERAG Inference}
\label{sec:BERAG Inference Latency}

In this section, we analyze the inference costs of BERAG and study acceleration techniques.

\paragraph{Compute} We first analyze asymptotic compute complexity for prefill and decode. The prefill time complexity is asymptotically $O(KD^2)$ for BERAG and $O(K^2D^2)$ for RAG given $K$ documents with length $D$ (Appendix \ref{app:inference_analysis}). This is because BERAG processes $K$ length-$D$ documents in parallel, whereas standard RAG processes one sequence of length $K \times D$. For decoding, BERAG requires the same asymptotic computation as RAG $O(K\times D)$, assuming all KV caches are available. However, in decoding BERAG pays an additional cost of $(K-1)\times P$ to attend to the shared prompt $\mathbf{x}$ (Eq.~\ref{eq:BERAG-probability}), where $P$ is the length of $\mathbf{x}$, and so we expect BERAG decoding to be slower than RAG. There are no savings in memory for  BERAG, as the KV caches for all documents still need to be stored.

\paragraph{Memory} We now discuss memory usage in  BERAG inference.
Let the vocabulary size be $V$.
The recursive form in Eq.~\ref{eq:berag posterior recursive form} requires us to store the previous posterior (K floats) and the next-token likelihood at the previous step ($V \times K$ floats). Given that V=150K for Qwen3-VL and assuming K=50, this requires 45 MB of memory per BERAG request.
In practice, we assign a buffer of 500 MB on each GPU. This allows our inference engine (vLLM-BERAG, see below) to handle ten K=50 user messages simultaneously. Once a message is processed, this space can be freed to be used for other requests.

\paragraph{Top-P Pruning Acceleration} We find a simple Top-P pruning procedure can substantially speed up BERAG decoding: at each decoding step, we keep the smallest number of documents such that their cumulative document posterior mass (Eq.\ref{eq:berag document posterior}) exceeds $P=1-\frac{1}{2K}$. This reduces compute and memory requirements for decode, as the KV cache associated with dropped documents no longer needs to be stored and operated on. We choose the threshold $P=1-\frac{1}{2K}$ as it naturally adjusts with respect to different $K$ settings and does not drop any document in the first pass given a uniform prior.

\begin{table}[!ht]
\centering
\small
\setlength{\tabcolsep}{3.5pt}
\renewcommand{\arraystretch}{1.12}
\begin{tabular}{@{}lccr@{}}
\toprule
\textbf{Method}
& \shortstack{\textbf{P90 TTFT ($\downarrow$)}\\\textbf{(s)}}
& \shortstack{\textbf{P90 TPOT ($\downarrow$)}\\\textbf{(ms)}}
& \shortstack{\textbf{VQA}\\\textbf{Score ($\uparrow$)}}\\
\midrule
\multicolumn{4}{c}{\textit{E-VQA} \quad (15.0K input tokens)} \\
\midrule
RAG
& 2.31~(1.00$\times$)
& 14.7~(1.00$\times$)
& 59.0 \\
BERAG
& {1.99~(0.86$\times$)}
& 19.5~(1.33$\times$)
& {72.3} \\
BERAG[TopP]
& 2.00~(0.86$\times$)
& {14.4~(0.98$\times$)}
& 72.1 \\
\midrule
\multicolumn{4}{c}{\textit{InfoSeek} \quad (94.9K input tokens)} \\
\midrule
RAG
& 35.2~(1.00$\times$)
& 25.2~(1.00$\times$)
& 36.3 \\
BERAG
& 10.5~(0.30$\times$)
& 35.7~(1.43$\times$)
& 42.4 \\
BERAG[TopP]
& {10.5~(0.30$\times$)}
& {17.2~(0.68$\times$)}
& {42.9} \\
\bottomrule
\end{tabular}
\caption{Inference efficiency and VQA performance at retrieval depth
$K=50$. TTFT denotes time to first token, and TPOT denotes time per
output token. Latencies are reported at the 90th percentile, with
times relative to RAG shown in parentheses. Values below
$1.0\times$ indicate faster inference. For BERAG[TopP], the pruning threshold is
$P=1-1/(2K)=0.99$.}
\label{tab:inference-efficiency}
\end{table}

\paragraph{Empirical validation with vLLM-BERAG} To validate our analysis and compare against standard RAG in a practical setup, we implement a custom version of vLLM~\citep{vllm-library} that enables BERAG inference to make use of common optimization techniques such as PagedAttention and iteration-based scheduling~\citep{vllm-library}. We call our engine vLLM-BERAG.

In Table~\ref{tab:inference-efficiency}, we show the 90-Percentile Time-To-First-Token (P90 TTFT) and the Time-Per-Output-Token (P90 TPOT) of the Qwen3-VL-8B-Instruct models on E-VQA and on Infoseek for $K=50$ with batch size 1. This simulates a latency-focused online serving scenario and covers input lengths from 3.1K to 94.9K tokens.

For prefill (P90 TTFT), standard  concatenative RAG is substantially slower than both plain BERAG and BERAG with Top-P pruning, both of which have similar prefill times, as expected.

For decoding speed (P90 TPOT), we find that plain BERAG is consistently slower than standard concatenative RAG, also as expected.
However, BERAG with Top-P pruning yields the same VQA performance as plain BERAG, but with substantial speed improvements, particularly on InfoSeek with its long passages.

These results show that  BERAG document posteriors can guide inference so that documents with very small posterior weights can be safely dropped without affecting answer generation. This is also an indirect confirmation of the quality of the learned document posterior.
Relative to concatenative RAG, BERAG with Top-P pruning offers reduced latency, improved decoding speed, along with BERAG's improved VQA performance.

\section{Additional Results and Discussion}

Due to space constraints, we present additional findings and provide pointers to appendices here.

\paragraph{BEFT Learns to Predict Retrieval Success}
We show simple decision rules to predict whether retrieval is successful (i.e., whether the gold document is contained in the Top-K list) with BEFT systems (Table~\ref{tab:beft_predict_retrieval_success}). We find introducing an empty pseudo-document and using its associated posterior distribution from BERAG gives accurate prediction. We note that this approach may serves as a more principled alternative to resolving external and parametric knowledge, as explored in recent VQA works~\citep{ye2026real, hong2026ccvqa}.

\begin{table}[ht]
\centering
\small
\setlength{\tabcolsep}{6pt}
\begin{tabular}{@{}c l cccc@{}}
\toprule
\# &  & \textbf{K=1} & \textbf{K=2} & \textbf{K=3} & \textbf{K=5} \\
\midrule
1 & GPT4o-mini & 77.3\% & 79.3\% & 79.3\% & 75.8\% \\
2 & Qwen2.5-VL & 70.6\% & 80.2\% & 78.5\% & 77.3\% \\ \midrule
3 & Qwen2-VL & 63.2\% & 67.2\% & 65.6\% & 64.8\% \\
4 & ~+ BEFT & 89.4\% & 82.8\% & 77.3\% & 72.3\% \\
5 & ~+ BEFT[w/$z_0$] & \textbf{93.2\%} & \textbf{90.0\%} & \textbf{86.8\%} & \textbf{83.3\%} \\
\bottomrule
\end{tabular}
\caption{Accuracy in predicting retrieval success or failure on 256 Infoseek examples. Line 1-3 show pre-trained models prompted to decide whether the Top-K documents are sufficient for question answering. BEFT uses prior-logits-based prediction. BEFT[w/$z_0$] uses posterior-based prediction, which flags retrieval failure when the pseudo passage $z_0$ has the highest posterior. Full description is in Appendix~\ref{app:predict-retrieval-success}.}
\label{tab:beft_predict_retrieval_success}
\vspace{-1em}
\end{table}

\paragraph{Experiments on Long Context VQA}

Motivated by BERAG's strong performance with long retrieval lists (Sec.~\ref{sec:exp-E-VQA and Infoseek Performance with respect to Context Length}) and invariance to document positions (Sec.~\ref{sec:gt-doc-position-analysis}), we evaluate BERAG on two established tasks with many items in context: SlideVQA (Sec.~\ref{sec:exp-SlideVQA Evidence Selection and Question Answering}) and MMNeedle (Sec.~\ref{sec:mmneedle}). We find that BEFT performs well on SlideVQA which requires multiple pieces of evidence for question answering, and that it achieves strong performance in multi-modal Needle-In-A-Haystack (NIAH) problems with minimal training.

\paragraph{Interpretability and Case Analysis}

We provide qualitative analyses on SlideVQA with quantitative measures of how the posterior distribution evolves during generation (Appendix~\ref{app:qualitative-analysis-case-slidevqa}). We find that, because the posterior distribution indicates which document the model was focusing on during inference, we are able to perform detailed error attribution.

\section{Conclusion}

We introduce Bayesian Ensemble Retrieval Augmented Generation (BERAG), a Bayesian ensemble-based RAG inference procedure along with Bayesian Ensemble Fine-Tuning (BEFT) to train BERAG systems end-to-end. We show BEFT-trained BERAG systems give strong performance on E-VQA and Infoseek by utilizing deep retrieval lists. We also demonstrate BERAG mitigates the ``lost-in-the-middle'' effect  via position-invariant ensembles. Furthermore, we show BERAG is very efficient in terms of inference. BERAG requires shorter prefill time than standard RAG by design, and can utilize document pruning to reduce decoding time without performance degradation. We show that  BERAG achieves substantially faster inference than standard RAG with long contexts in a modern inference framework such as vLLM. Overall, BERAG is a simple, fast, and good-performing approach to KB-VQA.

\section*{Limitations}

We formulate BEFT as a likelihood-based fine-tuning procedure.   We have not explored reinforcement learning based approaches.  However, we note that the BERAG framework provides a well-formulated probability distribution over output sequences (Eq.~\ref{eq:BERAG-probability}) that could be incorporated into RL for learning BERAG systems.
In this work, we focus on direct optimization of the likelihood (BEFT) and leave RL extensions to future work.

Another limitation is that we find that pre-trained systems require BEFT training to perform useful BERAG inference. This may limit BERAG to scenarios in which task-based fine-tuning is feasible, although we note that BERAG training is effective with lightweight methods such as LORA.

\bibliography{custom}

\appendix

\section{Illustration of BERAG}

\begin{figure*}[!ht]
    \centering
    \includegraphics[width=1.0\linewidth]{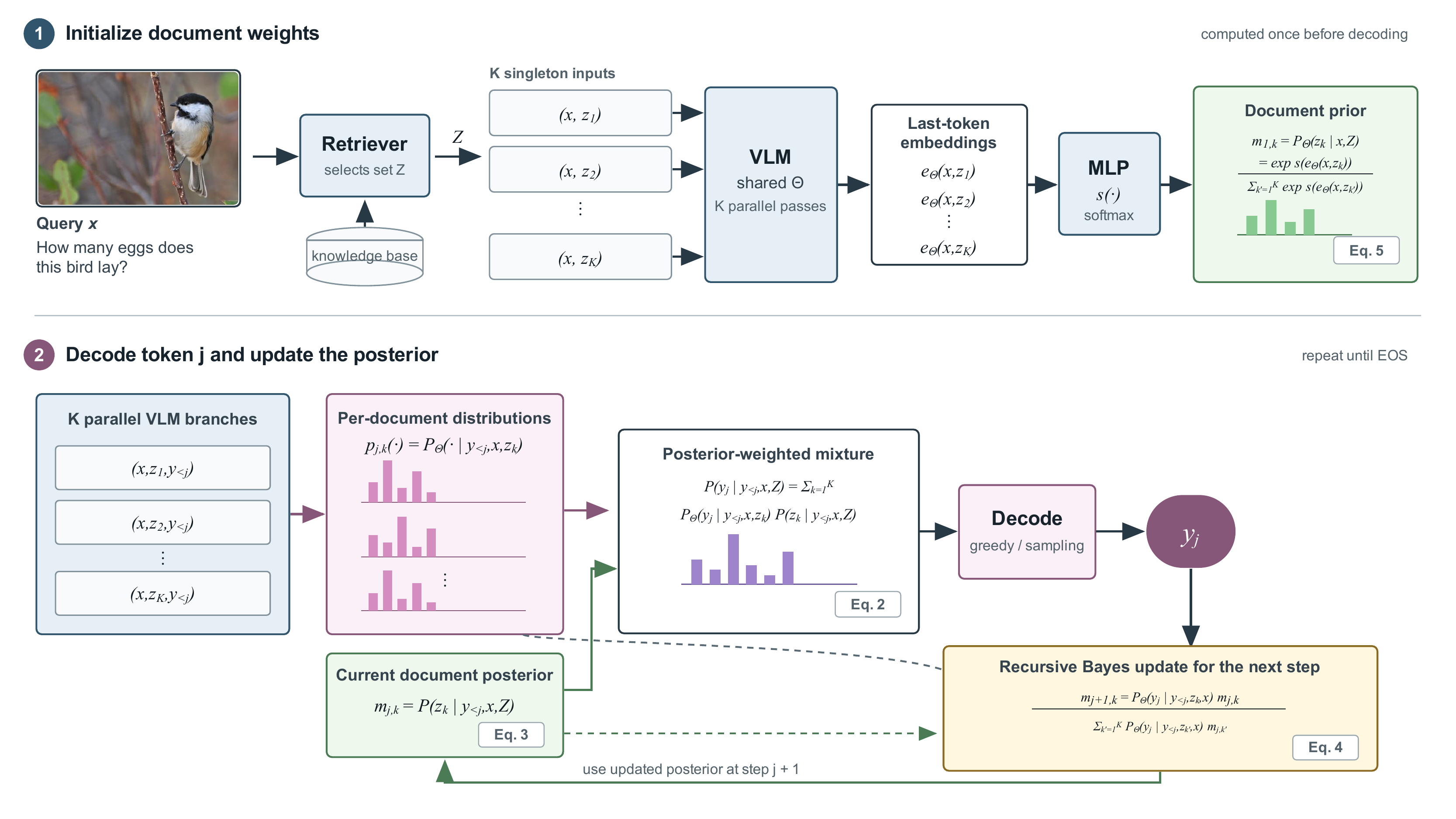}
    \caption{The inference procedure for Bayesian Ensemble Retrieval Augmented Generation (BERAG). Given query $x$ and $K$ documents, BERAG maintains $K$ independent branches of next-token likelihood computation, each conditioning on a document singleton $z_k$. These branches are weighted by the document posterior, which is computed via Bayes' rule using the document prior distribution and likelihood of previously generated tokens.}
    \label{fig:berag-system-diagram}
    \vspace{-1em}
\end{figure*}

Figure \ref{fig:berag-system-diagram} shows the inference procedure of BERAG.

\section{Extended Related Work}
\label{app:extended_related_work}

\paragraph{RAG with Document-level Ensembling} Aggregating generations conditioned on individual documents is considered in early RAG works~\citep{RAG-Lewis-NIPS2020} and continues to receive academic attention recently~\citep{Context-aware-decoding, entropy-based-decoding, RMCD}.
RMCD~\citep{RMCD} combines logits from multiple retrieved contexts using contrastive decoding, improving robustness to irrelevant contexts without additional training. Entropy-based decoding~\citep{entropy-based-decoding} proposes a similar document-parallel ensemble decoding scheme but uses entropy as heuristic signal to compute document weights. Compared to these works, we derive a principled probabilistic framework for inference and show an end-to-end training technique for the ensemble.

\paragraph{Long-Context Visual Question Answering}

Two main categories of tasks have emerged for testing VLM's answer generation capability on long, multi-modal context: \emph{Multi-modal Long Document QA}~\citep{mmlongbench-doc-nips2024, m-longdoc-emnlp2025, SlideVQA, dong2025mmdocrag} and \emph{Multi-modal Needle-in-a-Haystack}~\citep{mm-niah-nips2024, VisualHaystack-ICLR2025, MMNeedle-dataset}. For multi-modal long document QA, there is a notable trend in curating datasets with longer context and more confounders. SlideVQA~\citep{SlideVQA} provides 20 powerpoint slides as context for each question. MMDocRAG~\citep{dong2025mmdocrag} provides up to 20 quotes (i.e., text or image segments of a document) for each question with long-form answers. MMLongBench~\citep{mmlongbench-doc-nips2024} provides on average 47.5 pages for extractive question answering. M-LongDoc~\citep{m-longdoc-emnlp2025} provides on average 210.8 pages for long-form question answering. For multi-modal needle-in-a-haystack, MM-NIAH~\citep{mm-niah-nips2024} uses interleaving images and texts as haystack with scattered relevant information. Visual Haystack~\citep{VisualHaystack-ICLR2025} focuses on multi-image question answering and tests model ability to both retrieve images from a long context and to synthesize answers across them. MMNeedle~\citep{MMNeedle-dataset} composes stitched image panels as the haystack and focuses on retrieving the correct sub-image given the associated caption. In this work, we conduct experiments on SlideVQA and MMNeedle, two published benchmarks where training data is readily available. Given our focus on Knowledge-based Visual Question Answering, we leave evaluating BERAG and BEFT on other long-context VQA tasks to future work.

\paragraph{Tackling Lost-in-the-Middle Effect}

Lost-in-the-middle, a common failure mode of LLM-based response generation arising from overlooking pertinent information positioned in the middle of LLM's context, has been studied in depth in recent literature~\citep{liu-etal-2024-lost, mm-lost-in-the-middle-whosinspotlight, lost-in-multiturn-conversation}. The term was first coined by \citet{liu-etal-2024-lost} and recently studied in multi-modal contexts by \citet{mm-lost-in-the-middle-whosinspotlight} and in multi-turn dialogue by \citet{lost-in-multiturn-conversation}. These study report substantial performance degradation from the lost-in-the-middle and jointly identify the mitigation of this effect as a critical topic for further research.

A wide range of mitigation approaches have been proposed. Data-driven approaches attempt to alleviate lost-in-the-middle by curating a training distribution that emphasize response generation based on the middle of the context~\citep{In2Training-InformationIntensive-NIPS2024}. Another line of work focuses on long-context friendly schemes of positional encoding: CREAM~\citep{CREAM-NIPS2024} interpolates positional encoding by manipulating position indices; Ms-PoE~\citep{Ms-PoE-NIPS2024} introduces position-aware head-wise re-scaling ratios to native RoPE~\citep{RoPE-paper} encoding; and V2PE~\citep{V2PE-iccv2025} proposes a separate encoding scheme for visual tokens at variable and smaller increments. Another approach to tackle lost-in-the-middle is to subscribe to a response pipeline that is designed to handle long context. PAM QA~\citep{PAMQA-ACL2024} subscribes to a fixed pipeline of Question Repetition, Index Prediction, and finally Answer Summarization, whereas LongRAG~\citep{LongRAG-EMNLP2024} introduces a pipeline that includes CoT-guided filtering and LLM-based information extractor to tackle RAG tasks with long context.

Similar to previous works, we are also interested in establishing positional invariance. Notably, BERAG achieves this invariance without modifying positional encoding or stipulating task-specific response pipelines. To our knowledge, this is the first work that addresses lost-in-the-middle via positional-agnostic ensembles.

\section{Derivation for Recursive BERAG Inference}
\label{app:derivation}

In this section, we derive Eq. \ref{eq:berag posterior recursive form}.

{\small
\begin{align}
&P(z_k \mid y_{<j}, \mathbf{x}, Z) \notag
=
P(z_k \mid y_{j-1}, y_{<j-1}, \mathbf{x}, Z)
\notag \\
&= \notag \\
& \frac{
P(y_{j-1} \mid z_k, y_{<j-1}, \mathbf{x}, Z)
P(z_k \mid y_{<j-1}, \mathbf{x}, Z)
}{
P(y_{j-1} \mid y_{<j-1}, \mathbf{x}, Z)
}
\notag \\
&= \notag \\
&\frac{
P_\Theta(y_{j-1} \mid y_{<j-1}, z_k, \mathbf{x})
P(z_k \mid y_{<j-1}, \mathbf{x}, Z)
}{
\sum_{k'=1}^K
P_\Theta(y_{j-1} \mid y_{<j-1}, z_{k'}, \mathbf{x})
P(z_{k'} \mid y_{<j-1}, \mathbf{x}, Z)
} \notag
\end{align}
}

We use the convention that $y_{<0}=\emptyset$ and that $P(y_{<0})=1$. This ensures that Eq.~\ref{eq:berag posterior recursive form} is well-defined at the boundary.

\section{Supervised Learning of the Document Prior}
\label{sec:SLDP}
Optionally, we allow for an auxiliary classification loss to train the prior head $s(\cdot)$ with the document labels $r^{i}_{k}$. We interpret $\sigma(s_{k}^i)=\sigma(s(\mathbf{e}_\Theta(\mathbf{x}^i, z_k^i)))$ as the predicted probability that $z_k^i$ is relevant for answering query $\mathbf{x}^i$ where $\sigma(\cdot)$ denotes the sigmoid function. The auxiliary prior loss then follows from binary cross-entropy:

\begin{align}
        \mathcal{L}_{prior}=&-\frac{1}{NK}\sum_{i=1}^N\sum_{k=1}^K [r_k^i\log\sigma(s_{k}^i) \nonumber \\
        &+(1-r_k^i)\log(1-\sigma(s_{k}^i))]
    \label{eq:beft_prior_loss}
\end{align}

Unless specified, BEFT is done without $\mathcal{L}_{prior}$.

\section{BEFT Update Gradient Analysis}
\label{app:BEFT Update gradients}

In this section, we derive the gradients of BEFT loss. In the following derivation, we drop the superscript $i$ which indexes different training instances for readability. We also define the shorthands $p_{j,k} = P(y_j | y_{<j}, z_k, \mathbf{x})$ (Eq.~\ref{eq:BEFT-loss}) for the next-token probability at the $j^{th}$ position given the $k^{th}$ document and $m_{j,k} = P(z_k | y_{<j}, \mathbf{x}, Z)$ (Eq.~\ref{eq:berag document posterior}) for the document posterior probability for the $k^{th}$ document at the $j^{th}$ decoding step. These are defined for compactness. Let the sequence have $J$ tokens and let the collection $Z$ have $K$ documents.
{\small
\begin{align}\nabla_\Theta &\mathcal{L}_{BEFT}(\Theta)
    = -\log P(\mathbf{y} | \mathbf{x}, Z)
    \notag \\
    &= -\nabla_\Theta \sum_{j=1}^J
    \log \Bigl(
        \sum_{k=1}^K
        P(y_j, z_k | y_{<j}, \mathbf{x}, Z)
    \Bigr)
    \notag \\
    &= -\nabla_\Theta \sum_{j=1}^J
    \log \Biggl(
        \sum_{k=1}^K
        P_\Theta(y_j | y_{<j}, z_k, \mathbf{x})
        \notag \\
    &\hspace{5.5em}
        {}\times
        P_\Theta(z_k | y_{<j}, \mathbf{x}, Z)
    \Biggr)
    \notag \\
    &= -\nabla_\Theta \sum_{j=1}^J
    \log \Bigl(
        \sum_{k=1}^K p_{j,k} m_{j,k}
    \Bigr)
    \notag \\
    &= -\sum_{j=1}^J
    \frac{1}{
        \sum_{k'=1}^K p_{j,k'}m_{j,k'}
    }
    \nabla_\Theta
    \Bigl(
        \sum_{k=1}^K p_{j,k}m_{j,k}
    \Bigr)
    \notag \\
    &= -\sum_{j=1}^J
    \frac{1}{
        \sum_{k'=1}^K p_{j,k'}m_{j,k'}
    }
    \notag \\
    &\quad {}\times
    \Biggl(
        \sum_{k=1}^K
        \nabla_\Theta(p_{j,k})m_{j,k}
        +
        p_{j,k}\nabla_\Theta(m_{j,k})
    \Biggr)
    \notag \\
    &= -\sum_{j=1}^J\sum_{k=1}^K
    \Biggl[
        \frac{m_{j,k}}{
            \sum_{k'=1}^K p_{j,k'}m_{j,k'}
        }
        \nabla_\Theta(p_{j,k})
        \notag \\
    &\hspace{5.5em}
        {}+
        \frac{p_{j,k}}{
            \sum_{k'=1}^K p_{j,k'}m_{j,k'}
        }
        \nabla_\Theta(m_{j,k})
    \Biggr]
    \notag \\
    &= -\sum_{j=1}^J\sum_{k=1}^K
    \Biggl[
        \underbrace{
            \Delta_p(j,k)
            \nabla_\Theta
            \Bigl(
                P_\Theta(
                    y_j | y_{<j}, z_k, \mathbf{x}
                )
            \Bigr)
        }_{
            \substack{
                \text{weighted next-token}\\
                \text{likelihood gradient}
            }
        }
        \notag \\
    &\hspace{2.5em}
        {}+
        \underbrace{
            \Delta_m(j,k)
            \nabla_\Theta
            \Bigl(
                P_\Theta(
                    z_k | y_{<j}, \mathbf{x}, Z
                )
            \Bigr)
        }_{
            \substack{
                \text{weighted posterior}\\
                \text{gradient}
            }
        }
    \Biggr]
    \label{eq:beftgrad_over_k}
\end{align}
}

Eq.~\ref{eq:beftgrad_over_k} shows that the BEFT gradient can be decomposed into a token-level mixture of gradients for maximizing next-token likelihood given each document and for maximizing the document posterior. We have defined the gradient scale factors:

\begin{align}
\Delta_p(j,k)
    &= \frac{m_{j,k}}{
        \sum_{k'=1}^K p_{j,k'}m_{j,k'}
    }
    \notag \\
    &= \frac{
        P_\Theta(z_k | y_{<j}, \mathbf{x}, Z)
    }{
        \displaystyle
        \sum_{k'=1}^K
        \begin{aligned}[t]
        &P_\Theta(
            y_j | y_{<j}, z_{k'}, \mathbf{x}
        )\\[-2pt]
        &{}\times
        P_\Theta(
            z_{k'} | y_{<j}, \mathbf{x}, Z
        )
        \end{aligned}
    }
    \notag \\
    &= \frac{
        P_\Theta(z_k | y_{<j}, \mathbf{x}, Z)
    }{
        P_\Theta(y_j | y_{<j}, \mathbf{x}, Z)
    },
    \\
\Delta_m(j,k)
    &= \frac{p_{j,k}}{
        \sum_{k'=1}^K p_{j,k'}m_{j,k'}
    }
    \notag \\
    &= \frac{
        P_\Theta(y_j | y_{<j}, z_k, \mathbf{x})
    }{
        \displaystyle
        \sum_{k'=1}^K
        \begin{aligned}[t]
        &P_\Theta(
            y_j | y_{<j}, z_{k'}, \mathbf{x}
        )\\[-2pt]
        &{}\times
        P_\Theta(
            z_{k'} | y_{<j}, \mathbf{x}, Z
        )
        \end{aligned}
    }
    \notag \\
    &= \frac{
        P_\Theta(y_j | y_{<j}, z_k, \mathbf{x})
    }{
        P_\Theta(y_j | y_{<j}, \mathbf{x}, Z)
    }.
\end{align}

where $\Delta_p(j,k)$ scales the next-token likelihood gradient at position $j$ given document $z_k$. We name it the \emph{next-token likelihood gradient scale factor}.  $\Delta_m(j,k)$ scales the gradient for the posterior probability of $z_k$ at position $j$, given all previously generated tokens $y_{<j}$. We name it the \emph{document posterior gradient scale factor}. Note that $\Delta_p(j,k)$ and $\Delta_m(j,k)$ have a common denominator. These quantities will be useful for analysing the training dynamics of BEFT.

To see how BEFT scales the gradients for different documents, consider two documents $z_{k_1}$ and $z_{k_2}$ and their associated gradients at a certain position $j$. Consider the ratio between their next-token likelihood gradient scale factors:

\begin{align}
    \frac{\Delta_p(j,k_1)}{\Delta_p(j,k_2)} = \frac{P_\Theta(z_{k_1}|y_{<j}, \mathbf{x}, Z)}{P_\Theta(z_{k_2}|y_{<j}, \mathbf{x}, Z)} \label{eq:beft-next-token-likelihood-grad-ratio}
\end{align}
note that the common denominator cancels out. Eq.~\ref{eq:beft-next-token-likelihood-grad-ratio} shows that the posterior distribution under the current model modulates how aggressively BEFT maximizes the next-token likelihood given a document. Suppose $P_\Theta(z_{k_1}|y_{<j}, \mathbf{x}, Z) > P_\Theta(z_{k_2}|y_{<j}, \mathbf{x}, Z)$, then $\Delta_p(j,k_1) > \Delta_p(j,k_2)$ and so BEFT more aggressively maximizes the likelihood of $y_j$ given history $y_{<j}$ and the document $z_{k_1}$ compared to when the document $z_{k_2}$ is given (Eq.~\ref{eq:beftgrad_over_k}).

Now consider the ratio between their document posterior gradient scale factor:

\begin{align}
    \frac{\Delta_m(j,k_1)}{\Delta_m(j,k_2)} = \frac{P_\Theta(y_j|y_{<j},z_{k_1}, \mathbf{x})}{P_\Theta(y_j|y_{<j},z_{k_2}, \mathbf{x})} \label{eq:beft-posterior-grad-ratio}
\end{align}
similarly the common denominator cancels out. Eq.~\ref{eq:beft-posterior-grad-ratio} shows that the next-token likelihood modulates how BEFT maximizes the document posterior. If token $y_j$ is likely under document $z_{k_1}$ compared to document $z_{k_2}$, then BEFT will allocate more magnitude in gradients to maximize the posterior $P_\Theta(z_{k_1}|y_{<j}, \mathbf{x}, Z)$ for document $z_{k_1}$ compared to the posterior $P_\Theta(z_{k_2}|y_{<j}, \mathbf{x}, Z)$ for document $z_{k_2}$.

It is worth noting that, unlike contrastive learning, we do not specify which document is the ground-truth. Eq.~\ref{eq:beft-posterior-grad-ratio} shows that BEFT derives posterior learning signals from the model's own next-token likelihood. If we assume that the relevant document $z_{k^*}$ well-supports the ground-truth response, in the sense that the model assigns higher likelihood to the target token when provided with $z_{k^*}$ compared to when provided with other documents $z_{k\neq k^*}$, then BEFT training works to place high posterior probability on the relevant document $z_{k^*}$.

We will now show that, as a consequence of our Bayesian formalism, the gradient for document posterior $\nabla_\Theta P_\Theta(z_k | y_{<j}, \mathbf{x}, Z)$ is related to that of the document prior $\nabla_\Theta P_\Theta(z_k | \mathbf{x}, Z)$. Consider the document posterior gradient for document $z_k$ at position $j$:

\begin{align}
&\nabla_\Theta
\Bigl(
    P_\Theta(z_k | y_{<j}, \mathbf{x}, Z)
\Bigr)
\notag \\
&=
P_\Theta(z_k | y_{<j}, \mathbf{x}, Z)
\notag \\
&\quad {}\times
\nabla_\Theta
\Biggl(
    \log
    \Bigl(
        P_\Theta(z_k | y_{<j}, \mathbf{x}, Z)
    \Bigr)
\Biggr)
\notag \\
&=
P_\Theta(z_k | y_{<j}, \mathbf{x}, Z)
\notag \\
&\quad {}\times
\nabla_\Theta
\Biggl(
    \log
    \Biggl(
        \frac{
            \begin{aligned}
            &P_\Theta(y_{<j} | z_k, \mathbf{x})\\[-2pt]
            &{}\times P_\Theta(z_k | \mathbf{x}, Z)
            \end{aligned}
        }{
            P(y_{<j} | \mathbf{x}, Z)
        }
    \Biggr)
\Biggr)
\notag \\
&\propto
\frac{
    P_\Theta(z_k | y_{<j}, \mathbf{x}, Z)
}{
    P_\Theta(z_k | \mathbf{x}, Z)
}
\nabla_\Theta
P_\Theta(z_k | \mathbf{x}, Z)
\label{eq:posterior_grad_decomposition}
\end{align}

Eq.~\ref{eq:posterior_grad_decomposition} shows that to maximize the posterior for document $z_k$, BEFT works to increases the prior probability placed on $z_k$. As we have shown with Eq.~\ref{eq:beft-posterior-grad-ratio} that the \emph{posterior} of a document singleton that well-explains the response receives larger-magnitude gradients, it follows that BEFT learns to place higher \emph{prior} probability on relevant document singletons. This can be regarded as a form of self-supervision that learns to perform evidence selection from question answering training. For KB-VQA and Long-context VQA tasks this is particularly useful, as the main challenge lies in identifying relevant information from a noisy context.

\section{Experimental Setup Details}
\label{app:experimental_setup_details}

\paragraph{Training Details and Hyper-parameters} For Infoseek, we subsample 60K examples from the training set for SFT, DPO, and BEFT training. For E-VQA, we use the full training set. We use a LoRA rank of 64 in training with a learning rate of 1e-5.

\paragraph{Inference} We use FP16 weights and greedy decoding. We find that temperature sampling gives similar performance and so have kept to deterministic decoding.

\paragraph{Libraries} We use LLaMAFactory~\citep{zheng2024llamafactory} to perform supervised fine-tuning and DPO training, and implement a custom trainer in LLaMAFactory for BEFT training. We use vLLM for SFT and DPO inference, and implement BERAG inference with the transformers~\citep{transformers-library} library and our custom vLLM-BERAG engine.

\paragraph{MLP head for document prior} We use a two-layer MLP head $s(\cdot)$ to compute the document prior (Eq.\ref{eq:berag document prior}). The MLP has the same hidden dimensions as the backbone vision language model. For Qwen2-VL-Instruct, we use the `<im\_end>` embedding. For LLaVA-LLama model, we use the `<eot\_id>` embedding. In training, we apply a small learning rate of 1e-7 to fine-tune the MLP. We find that using a larger learning rate (e.g., 1e-5 as applied to the backbone VLM) will result in unstable training and worse performance.

\paragraph{DPO $\beta$ values} On E-VQA we use $\beta=0.7$ and find that performance is insensitive to $\beta \in (0.3,0.9)$. On Infoseek, we find that a large value of $\beta=1.5$ is required to see gains from DPO. This heavy regularization may be attributed to the fact that Infoseek is curated based on very few Wikipedia documents compared to its training set size, which results in overfitting.

\section{Using BEFT model as a reranker}

In Table \ref{tab:retrieval_performance_evqa_infoseek_slidevqa}, we use the trained BEFT model's document prior distribution to perform reranking on the Top-50 document retrieved by PreFLMR-L on E-VQA and Infoseek. We find that the BEFT document prior-based reranker substantially improve Recall@K compared to the base retriever, and note that this is achieved with minimal compute overhead.

\begin{table}[!t]
\centering
\small
\setlength{\tabcolsep}{5.5pt}
\begin{tabular}{lcccc}
\toprule
 & \textbf{R@1} & \textbf{R@3} & \textbf{R@5} & \textbf{R@10} \\
\midrule
\multicolumn{5}{c}{\textit{E-VQA}} \\ \midrule
MUKA & - & - & 64.9 & - \\
\midrule
PreFLMR-L & 39.5 & 56.6 & 60.6 & 68.0 \\
~~w/ BEFT & 59.7 & 71.5 & 75.4 & 79.6 \\
~~w/ BEFT[w/ $\mathcal{L}_{prior}$] & \textit{60.6} & \textit{72.2} & \textit{76.3} & \textit{80.3} \\
\midrule
\multicolumn{5}{c}{\textit{Infoseek}} \\
\midrule
Eva-CLIP-8B & 45.6 & - & 67.1 & 73.0 \\
\midrule
PreFLMR-L & 13.1 & 28.2 & 36.6 & 49.0 \\
~~w/ BEFT & 51.2 & \textit{66.0} & 69.5 & 73.0 \\
~~w/ BEFT[w/ $\mathcal{L}_{prior}$] & \textit{52.6} & \textbf{69.5} & 70.6 & 73.4 \\
\bottomrule
\end{tabular}
\caption{Recall@K (R@K) on E-VQA and Infoseek. Initial retrieval is done by Eva-CLIP or by PreFLMR-L. Reranker is denoted after ``w/''. With PreFLMR-L, reranking is performed on the Top-50 documents retrieved. BEFT refers to using the logits produced by the trained prior head for reranking.}
\label{tab:retrieval_performance_evqa_infoseek_slidevqa}
\end{table}

\paragraph{Adding auxiliary prior loss further improves BEFT-based reranking.} In Sec.\ref{sec:method}, we describe how to incorporate auxiliary classification loss $\mathcal{L}_{prior}$for training the prior. We find that adding $\mathcal{L}_{prior}$ gives stronger reranker. But we find that the system performs slightly worse on end-to-end VQA.

\section{Learning to Predict Retrieval Success}
\label{sec:abstention}
\label{app:predict-retrieval-success}

It can be useful for the answer generation system to predict whether the retrieval is successful, that is, whether the ground-truth document is returned in the Top-$K$ list. We now show two decision rules to predict retrieval success with BEFT systems: (1) \emph{Prior-logits-based}. We compute the relevance probabilities of all individual documents following the logistic classifier $\sigma(s_k^i)$ in Eq.\ref{eq:beft_prior_loss}. If any value exceeds $0.5$, we flag a retrieval success. This decision rule can be used directly by BEFT systems trained according to Sec.\ref{sec:experimental setup}; (2) \emph{Posterior-based}. We include an empty pseudo-passage $z_0$. A retrieval failure is flagged if the posterior (Eq.~\ref{eq:berag document posterior}) assigns the highest mass on $z_0$. This decision rule is applied to a variant of BEFT system, where in training we replace the gold document with an empty pseudo-passage $z_0$ on 50\% of the examples. We apply the prior loss in training (Eq.~\ref{eq:beft_prior_loss}) to encourage the model to assign high posterior to $z_0$ on retrieval failure. We refer to system trained with this data as BEFT[w/$z_0$].

\begin{table}[ht]
\centering
\small
\setlength{\tabcolsep}{6pt}
\begin{tabular}{@{}c l cccc@{}}
\toprule
\# &  & \textbf{K=1} & \textbf{K=2} & \textbf{K=3} & \textbf{K=5} \\
\midrule
1 & GPT4o-mini & 77.3\% & 79.3\% & 79.3\% & 75.8\% \\
2 & Qwen2.5-VL & 70.6\% & 80.2\% & 78.5\% & 77.3\% \\ \midrule
3 & Qwen2-VL & 63.2\% & 67.2\% & 65.6\% & 64.8\% \\
4 & ~+ BEFT & 89.4\% & 82.8\% & 77.3\% & 72.3\% \\
5 & ~+ BEFT[w/$z_0$] & \textbf{93.2\%} & \textbf{90.0\%} & \textbf{86.8\%} & \textbf{83.3\%} \\
\bottomrule
\end{tabular}
\caption{Accuracy in predicting retrieval success or failure on 256 Infoseek examples. Rows 1-3 show pre-trained models prompted to decide whether the Top-K documents are sufficient for question answering. BEFT uses prior-logits-based prediction. BEFT[w/$z_0$] uses posterior-based prediction, which flags retrieval failure when the pseudo passage $z_0$ has the highest posterior.}
\label{tab:beft_predict_retrieval_success-app}
\vspace{-1em}
\end{table}

We experiment on Infoseek using the PreFLMR-L retriever where recall is poor at small $K$. Table \ref{tab:beft_predict_retrieval_success-app} shows the accuracy for predicting retrieval success. Both BEFT and BEFT[w/$z_0$] under their respective decision rules perform well on the task compared to strong baselines.
In comparison, BEFT[w/$z_0$] outperforms standard BEFT by 4-11\% in accuracy, with gains widening at higher $K$. This shows that the posterior distribution in BERAG can be utilized for effective retrieval success prediction without introducing an additional prediction step. We note that these prediction results can be used to perform deflection (i.e., refusing to answer given insufficient context). We study this in Appendix \ref{app:deflection}.

\

\section{Results on Deflection}
\label{app:deflection}

\begin{table}[ht]
\centering
\small
\setlength{\tabcolsep}{6pt}
\begin{tabular}{@{}c l cccc@{}}
\toprule
\# &  & \textbf{K=1} & \textbf{K=2} & \textbf{K=3} & \textbf{K=5} \\
\midrule
\multicolumn{6}{c}{\textit{Deflection Accuracy}} \\
\midrule
1 & BEFT & 89.4\% & 82.8\% & 77.3\% & 72.3\% \\
2 & BEFT[w/$z_0$] & \textbf{93.2\%} & \textbf{90.0\%} & \textbf{86.8\%} & \textbf{83.3\%} \\
\midrule
\multicolumn{6}{c}{\textit{Deflection F1}} \\
\midrule
3 & BEFT & 0.93 & 0.88 & 0.82 & 0.74 \\
4 & BEFT[w/$z_0$] & \textbf{0.96} & \textbf{0.94} & \textbf{0.91} & \textbf{0.87} \\
\midrule
\multicolumn{6}{c}{\textit{VQA Score (Strict RAG)}} \\
\midrule
5 & All Deflect & 86.9 & 78.6 & 63.4 & 51.0 \\
6 & BEFT & 84.0 & 75.8 & 68.0 & 57.8 \\
7 & BEFT[w/$z_0$] & \textbf{89.2} & \textbf{83.5} & \textbf{78.1} & \textbf{72.0} \\
\bottomrule
\end{tabular}
\caption{Deflection accuracy, deflection F1 score, and VQA score under Strict RAG setting across different Top-$K$ values on Infoseek. For BEFT systems, deflection is issued when $z_0$ has the highest posterior (Eq.\ref{eq:berag document posterior}), not via prompting.
In Strict RAG, the system is marked as correct if it chooses to deflect when the ground-truth document is not in the retrieval list, so that ``All Deflect'' has a score $1-$Recall@K.}
\label{tab:beft_deflection}
\vspace{-1em}
\end{table}

We experiment on Infoseek where retrieval performance is poor at small $K$. Table \ref{tab:beft_deflection} shows the accuracy of deflection decision and F1 score for $K=1,2,3,5$, where we stipulate that a deflection should be flagged when the gold document is not present in the retrieved set. We find that BEFT[w/$z_0$] achieves substantially higher deflection accuracy and F1 compared to the BEFT under these Top-$K$ settings (Table \ref{tab:beft_deflection}, Lines 1-4).

We also evaluate these systems under the \textit{Strict RAG} setting. In Strict RAG, when the gold document is absent, the system is correct if and only if it deflects; otherwise, the system is correct if and only if the model does not deflect and does generates the correct answer. We find that BEFT[w/$z_0$] consistently outperforms BEFT under Strict RAG.

\section{Asymptotic Analysis for BERAG inference}
\label{app:inference_analysis}

In the analysis below we assume the standard, quadratic attention mechanism for language generation. We note that BERAG also enjoys improvement from any sub-quadratic version of attention mechanism in that each branch in the ensemble be computed more efficiently. We assume we are given $K$ documents each with length $D$.

\paragraph{Decode Analysis} In decoding, the incoming embedding interacts with all previous tokens at every layer, that is, the amount of compute required is determined by the length of KV cache. For plain BERAG, the length of KV cache is strictly larger than that in standard RAG. This is because the BERAG ensemble still need to store all document tokens as KV cache. In addition, query tokens need to be repeated $K$ times for each branch in the ensemble. The total KV cache is therefore strictly larger than that in standard RAG due to the repeated query tokens.

For BERAG with Top-P pruning, KV cache associated with low-posterior documents can be dropped, thereby leading to substantial speed-up in decoding.

\paragraph{Prefill Analysis}  The asymptotic compute complexity for standard RAG prefill is $O((KD)^2)=O(K^2D^2)$. This is because all documents are concatenated to form one context with length $KD$. For BERAG prefill, the asymptotic compute complexity is $O(KD^2)$, as each branch contains a length-$D$ context and there are $K$ branches in total to cover all documents. In this analysis, we have assumed that the query length $Q$ is negligible compared to the document length $D$. We note that the query is identical for every branch, and so a possible optimization is to encode the query once and reuse the associated KV cache for every branch. We leave such practical optimization techniques to future work and focus on quantifying the asymptotic behavior here.

\section{Experiments on Long Context VQA}
\label{sec:Experiments on Long Context VQA}

Motivated by BERAG's strong performance with long retrieval list (Sec.~\ref{sec:exp-E-VQA and Infoseek Performance with respect to Context Length}) and invariance to document positions (Sec.~\ref{sec:gt-doc-position-analysis}), we evaluate BERAG on two established tasks with many items in context: SlideVQA (Sec.~\ref{sec:exp-SlideVQA Evidence Selection and Question Answering}) and MMNeedle (Sec.~\ref{sec:mmneedle}). We also study inference optimization for BERAG which becomes critical with long context (Sec.~\ref{sec:BERAG Inference Latency}).

\subsection{Experimental Setup}

\paragraph{SlideVQA} The dataset consists of 10.6K and 2.22K questions for training and testing, respectively. The system is presented with a 20-slide presentation to answer each question. There can be multiple relevant slides for each question and ground-truth slides annotations are provided. Evaluations are by exact match for evidence selection and question answering following \citep{SlideVQA}. We perform BEFT with $K=4$.

\begin{figure*}[!htbp]
    \centering
    \includegraphics[width=0.9\textwidth]{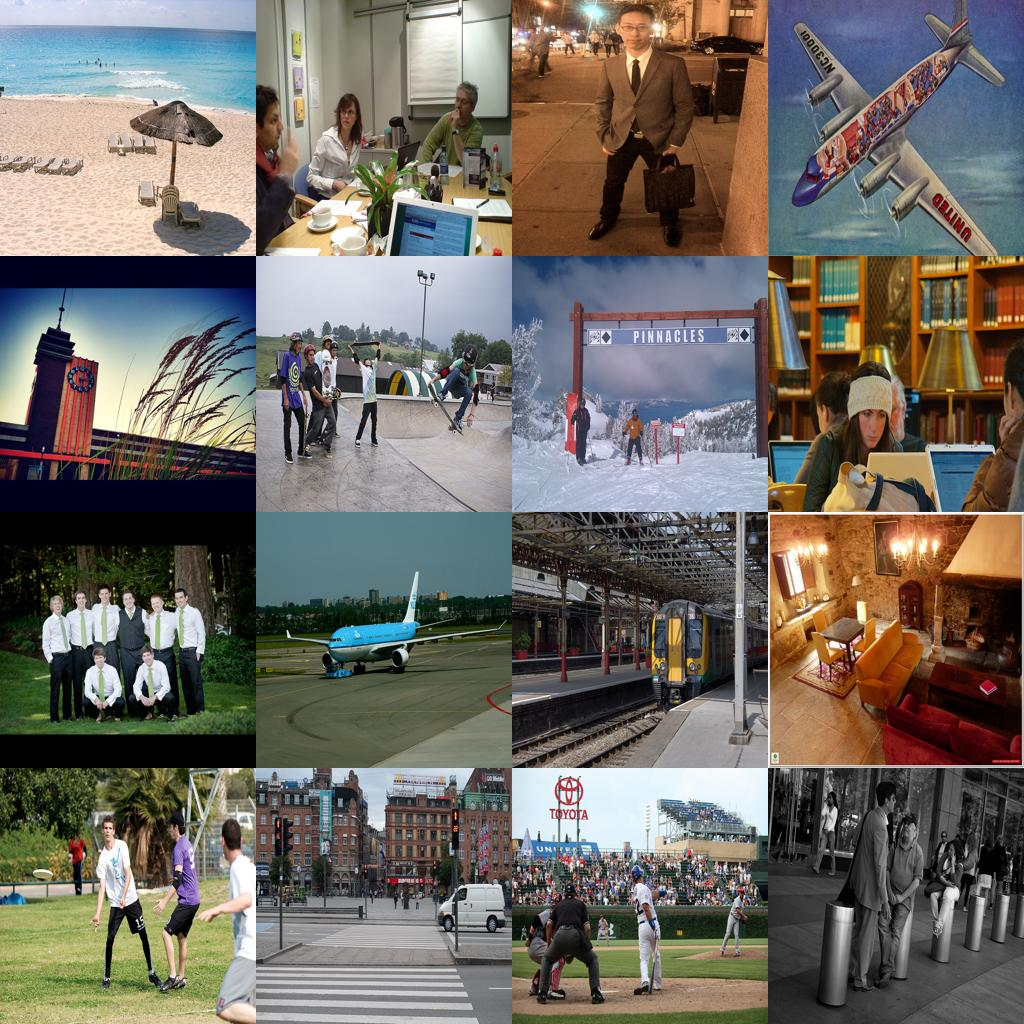}
    \caption{An example image panel from the MMNeedle benchmark for $4\times 4$ image panel (i.e., $N=4$). Assuming that this is the first image panel in the provided context and given the query ``A blue commercial airliner is parked on the airport tarmac near a terminal area.'', the model is expected to generate ``1, 3, 2''.}
    \label{fig:mmneedle-example}
\end{figure*}

\paragraph{MMNeedle} We use the setting where each haystack consists of $M=10$ panels. Each panel is formed by $N \times N$ sub-images (example in Figure \ref{fig:mmneedle-example}). We evaluate with the ``Exact'' metric in \citet{MMNeedle-dataset}: the model is given a caption to a sub-image and tasked to generate the index of the panel, and the column-row coordinate of the corresponding sub-image in the format of ``index, row, col''. Models should generate ``-1'' if the sub-image is not present. We perform BEFT with $K=2$ on LLava-Llama-3-8B on training data curated from COCO2014 training split.

\subsection{SlideVQA Evidence Selection and Question Answering}
\label{sec:exp-SlideVQA Evidence Selection and Question Answering}

We study SlideVQA to understand if BERAG can utilize multiple, multi-modal documents.
Table \ref{tab:slidevqa_combined} reports Evidence Selection (ES) and Question Answering (QA) performance on SlideVQA.

\begin{table}[!ht]
\centering
\small
\begin{tabular}{lcc}
\toprule
\textbf{Method} & \textbf{ES EM} & \textbf{QA EM} \\
\midrule
BM25                 & 35.9 & --   \\
DSE~\citep{DSE} & 67.2 & --   \\
VDocRAG~\citep{VDocRAG} & 73.3 & 44.2$^*$ \\
M3D~\citep{SlideVQA}                  & 75.0 & 33.5 \\
AVIR~\citep{avir}            & --   & 60.3 \\
Eagle-2.5~\citep{eagle2.5} & -- & 63.2 \\
 \midrule
Qwen2-VL-7B-Instruct & --   & 28.5 \\
~~+ SFT              & --   & 31.0 \\
~~+ BEFT (ours)               & 90.4  & \textbf{69.6} \\ \midrule
Human                & 97.7 & 89.8 \\
\bottomrule
\end{tabular}
\caption{Results on SlideVQA. Evidence Selection EM (ES EM) measures evidence selection accuracy, and QA Score is measured with Exact Match (QA EM). $^*$ denotes Token F1 as QA EM is not reported. Our BEFT system scores 76.5 under F1.}
\label{tab:slidevqa_combined}
\end{table}

\paragraph{BERAG and SoTA} SoTA systems follow two major approaches (1) \textit{Direct Generation}. Eagle-2.5~\citep{eagle2.5} directly generates the answer based on the 20-slide context and does not involve a retrieval or evidence selection step. Performance depends on the model's long-context capability; (2) \textit{Select-then-Generate}. M3D~\citep{SlideVQA}, VDocRAG~\citep{VDocRAG}, and AVIR~\citep{avir} first perform Evidence Selection (ES) with a retriever or a separate ES module followed by Question Answering (QA).

\paragraph{BERAG is a Direct Generation method with interpretable Evidence Selection.} Slides are directly provided to the system for generation  without pre-processing and the document posterior (Eq.~\ref{eq:berag document posterior}) provides an ordering of slides as they contribute to generation.
Our BEFT model achieves an Exact Match score of 90.4 and 69.6 on ES and QA, representing a $+15.4$ and $+6.4$ gains compared to recent baselines, respectively.

\paragraph{BERAG generalizes to multi-modal context with multiple relevant documents.}  SlideVQA contains instances where multiple slides are labeled as ground-truth. This is unlike E-VQA and Infoseek, which have a single ground-truth document.  The strong performance on SlideVQA shows that BEFT  can synthesize information from multiple multi-modal documents in context. We provide qualitative analysis in Appendix \ref{app:qualitative-analysis-case-slidevqa}.

\subsection{Multi-modal Needle-In-A-Haystack (NIAH) Performance}
\label{sec:mmneedle}

\begin{table}[!t]
\centering
\small
\setlength{\tabcolsep}{8pt}
\begin{tabular}{lcccc}
\toprule
\textbf{Model} & \textbf{1$\times$1} & \textbf{2$\times$2} & \textbf{4$\times$4} & \textbf{8$\times$8} \\
\midrule
Claude 3 Opus   & 66.9 & 4.6  & 0.4 & 0.0 \\
GPT-4o          & 97.0 & 81.8 & 26.9 & 1.0 \\ \midrule
LLaVA-Llama-3-8B   & 0.0  & 0.0  & 0.0 & 0.0 \\
\quad + BEFT (ours)   & 97.1   & 86.8   & 41.4 & 0.0  \\
\multicolumn{4}{l}{\quad w/ each $8\times 8$ maps to four $4\times 4$ s.t. M=40} & 42.5 \\
\bottomrule
\end{tabular}
\caption{Accuracy for the ``Exact'' metric on MMNeedle with $M=10$ images as the Haystack.  Results for Claude 3 Opus, GPT-4o, and LLaVA-Llama-3-8B are taken from \citet{MMNeedle-dataset}.}
\label{tab:mmneedle_performance}
\vspace{-1.5em}
\end{table}

We now show BEFT training followed by BERAG decoding is a competitive approach to Multi-Modal Needle-In-A-Haystack (NIAH) problems. Table \ref{tab:mmneedle_performance} shows that the BEFT-trained LLaVA-Llama-3-8B model outperforms GPT-4o and Claude 3 Opus with $M=10$ image panels where panels consists of $1\times 1, 2\times 2$ and $4\times 4$ sub-images (i.e.,  $N=1,2,4$). All models perform at near-zero performance for $N=8$ ($8\times 8$). This is likely due to limitations in the vision encoder. We note that it is possible to break each $8\times 8$ panel into four $4\times 4$ panels and conduct BERAG inference effectively at $M=40$. This gives comparable performance to the $4\times 4$ setup and compensates for limitations in the vision encoder.

\begin{table}[!ht]
\centering
\small
\setlength{\tabcolsep}{4.5pt}
\begin{tabular}{@{}l@{\hspace{4pt}}cc|cc|cc@{}}
\toprule
& \multicolumn{2}{c|}{$K=10$}
& \multicolumn{2}{c|}{$K=30$}
& \multicolumn{2}{c}{$K=50$} \\
& ms/tok & BEM
& ms/tok & BEM
& ms/tok & BEM \\
\midrule
RAG & 68.5  & 64.8 & 97.5  & 60.2 & 203.0 & 56.3 \\
BERAG & 95.0  & \textbf{69.7} & 279.2 & 71.5 & 470.2 & 70.3 \\
~~w/\textit{Top-P} & \textbf{44.8} & 69.1 & \textbf{45.6} & \textbf{71.9} & \textbf{44.4} & \textbf{70.7} \\
\bottomrule
\end{tabular}
\caption{Decoding speed (millsecond per token, ms/tok) and BEM score (Sec.~\ref{sec:experimental setup}) on 512 E-VQA examples under $K=10,30,50$. The \textit{Top-P} strategy prunes branches in BERAG based on the document posterior.}
\label{tab:decode_speed_vqa_comparison}
\end{table}

\section{Use of AI Assistants}

AI coding assistants are primarily used to (1) compose bash scripts from existing, manually-written templates; (2) migrate BEFT implementation to a newer version of LlamaFactory based on a manually written implementation for training newer models.

AI assistance was used to format LaTeX tables and prepare the arXiv version, including adding the project-page footnote and making minor editorial and typesetting corrections. All citations and references were manually added.

\section{Use and License of Artifacts}

We confirm we meet the licensing requirement of all artifact used. This includes the datasets (E-VQA~\citep{EVQA-dataset}, Infoseek~\citep{Infoseek-dataset}, SlideVQA~\citep{SlideVQA}, and MMNeedle~\citep{MMNeedle-dataset}), models (Qwen2-VL-Instruct~\citep{qwen2vl-technical-report} and Llava-Llama-3~\citep{llava-llama-model}), and libraries (Transformer~\citep{transformers-library}, LlamaFactory~\citep{zheng2024llamafactory}, vllm~\citep{vllm-library}). We will publish BEFT training and inference code under the MIT license.

\section{Qualitative Case Analysis on SlideVQA}
\label{app:qualitative-analysis-case-slidevqa}

In this section, we conduct qualitative analysis on successful and failed cases on SlideVQA. In each diagram, we show the question, the ground-truth slide(s), the ground-truth anser, the BEFT model's response, and the document posterior over generation as a heat map, with the x-ticks being token indices and the y-ticks being the ground-truth slide(s) and the other slides. We have aggregate the probability mass on all other slides except for the ground-truth into one y-tick labeled ``Other Documents''.

\begin{figure*}[!ht]
    \centering
    \includegraphics[width=1.0\linewidth]{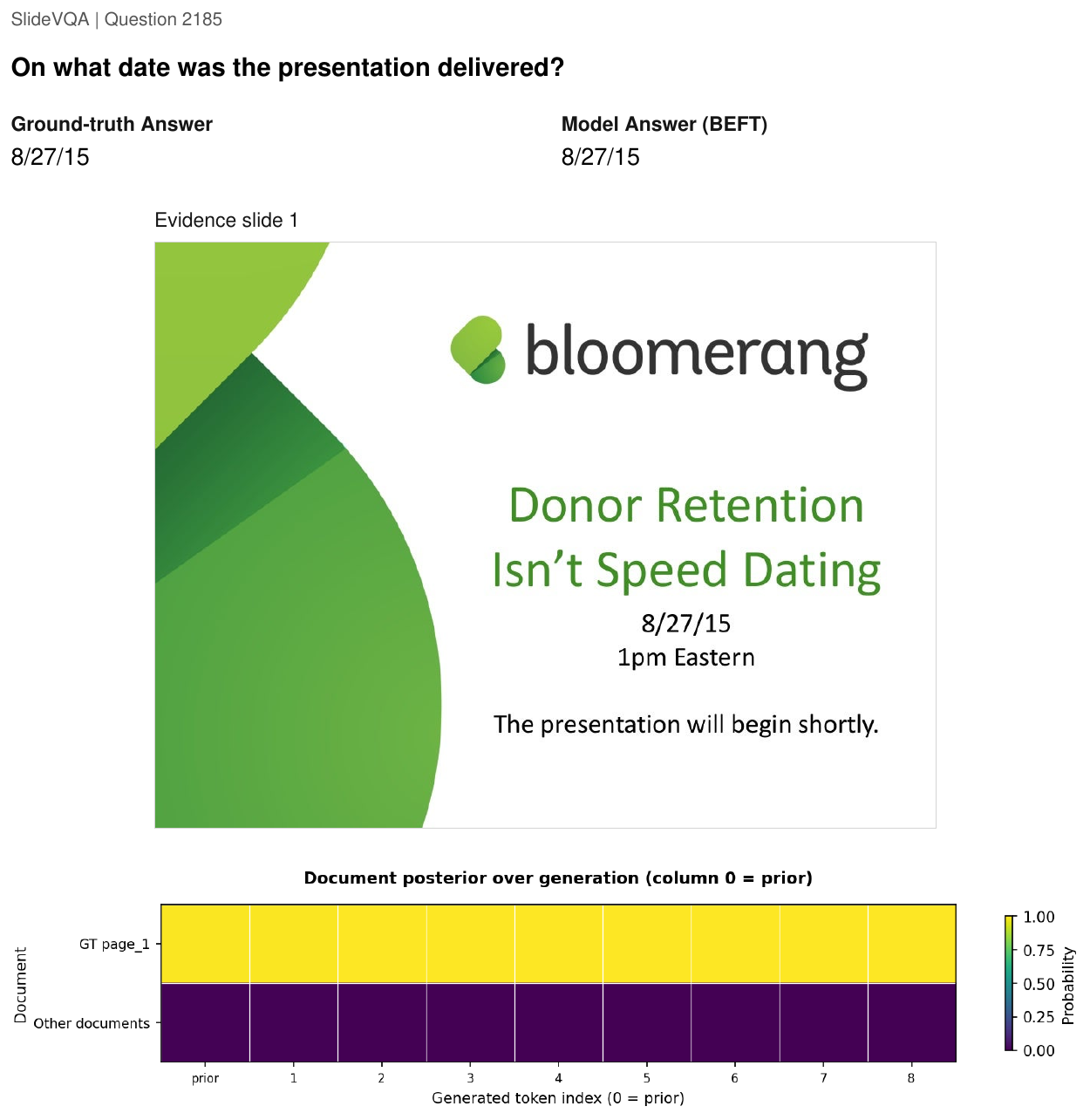}
    \caption{Question ID 2185. Correct answer with single ground-truth slide.}
    \label{fig:2185}
\end{figure*}

Figure \ref{fig:2185} shows a case where the model answers correctly. The model focuses on the correct slide throughout generation.

\begin{figure*}[!ht]
    \centering
    \includegraphics[width=1.0\linewidth]{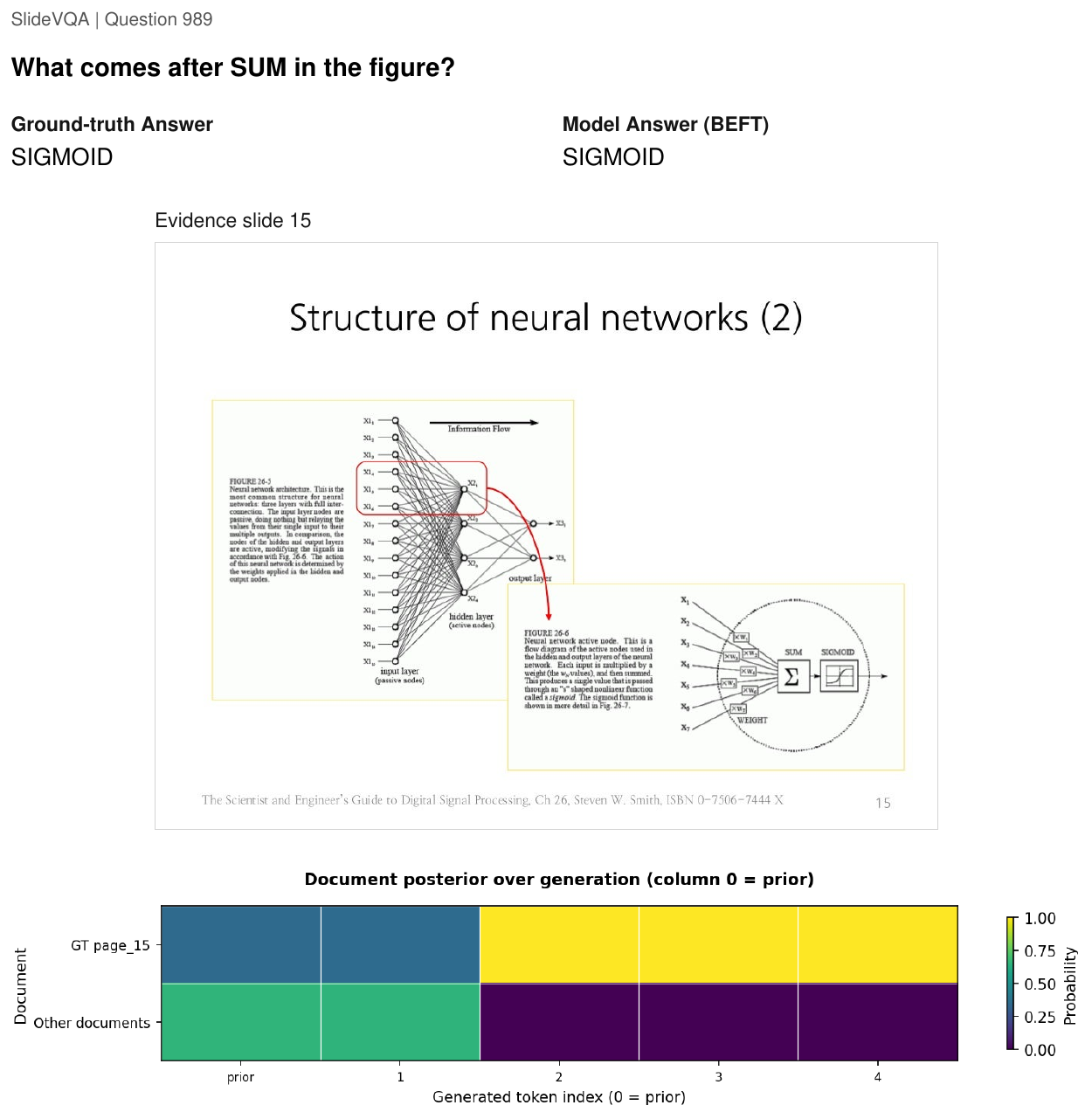}
    \caption{Question ID 989. Correct answer with single ground-truth slide. Posterior becomes more focused on the ground-truth slide at later tokens.}
    \label{fig:989}
\end{figure*}

Figure \ref{fig:989} also shows a correct case. The posterior is correctly placed on the ground-truth slide only at the end of generation, although the prior distribution is less certain.

\begin{figure*}[!ht]
    \centering
    \includegraphics[width=1.0\linewidth]{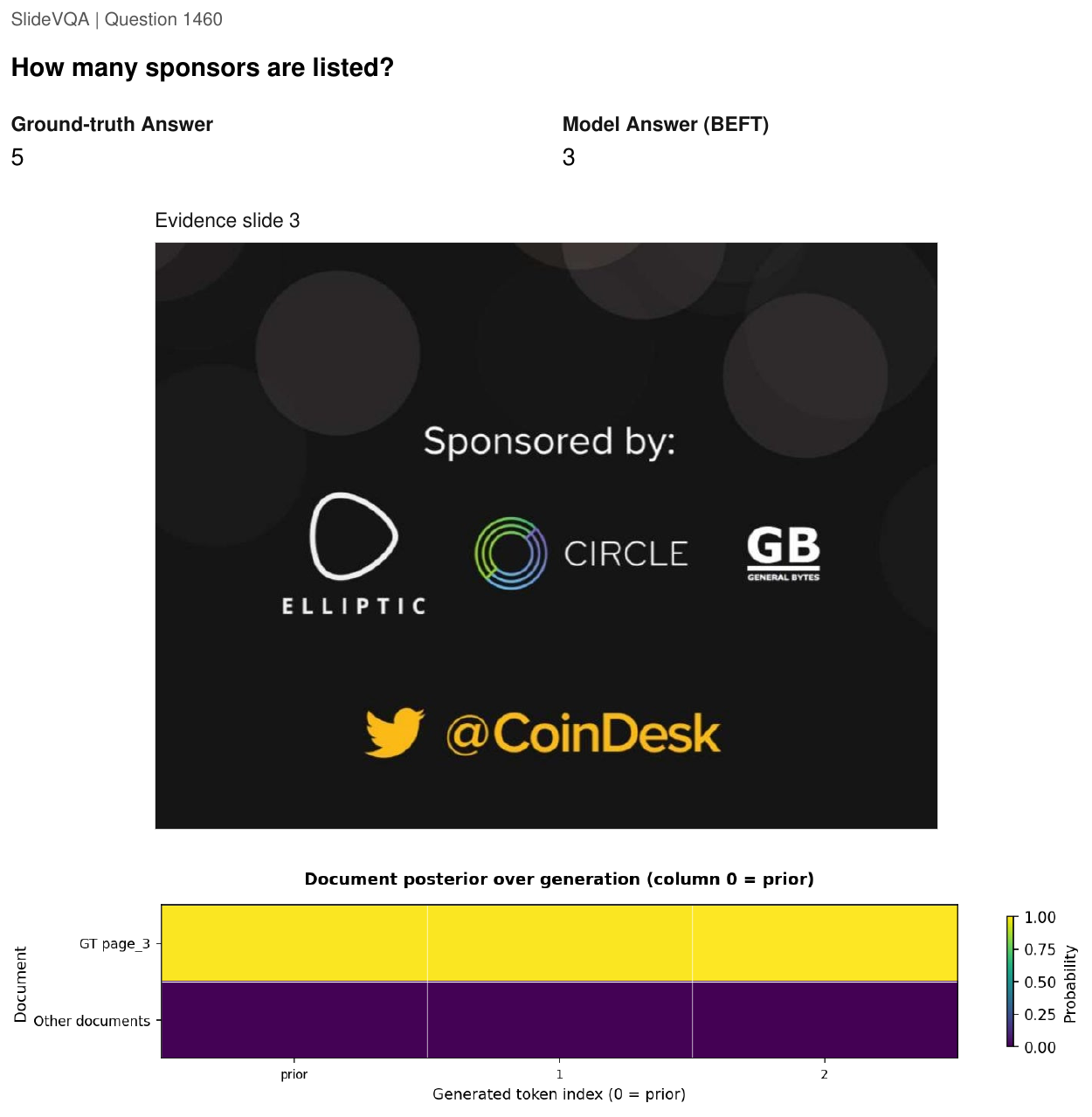}
    \caption{Question ID 1460. Wrong answer with single ground-truth slide, likely due to a counting failure.}
    \label{fig:1460}
\end{figure*}

Figure \ref{fig:1460} shows a failure case. The prior is correctly placed on the ground-truth slide, but the model likely miscounts the number of logos present. Interestingly, the concentrated document posterior enables us to accurately attribute error to counting.

\begin{figure*}[!ht]
    \centering
    \includegraphics[width=1.0\linewidth]{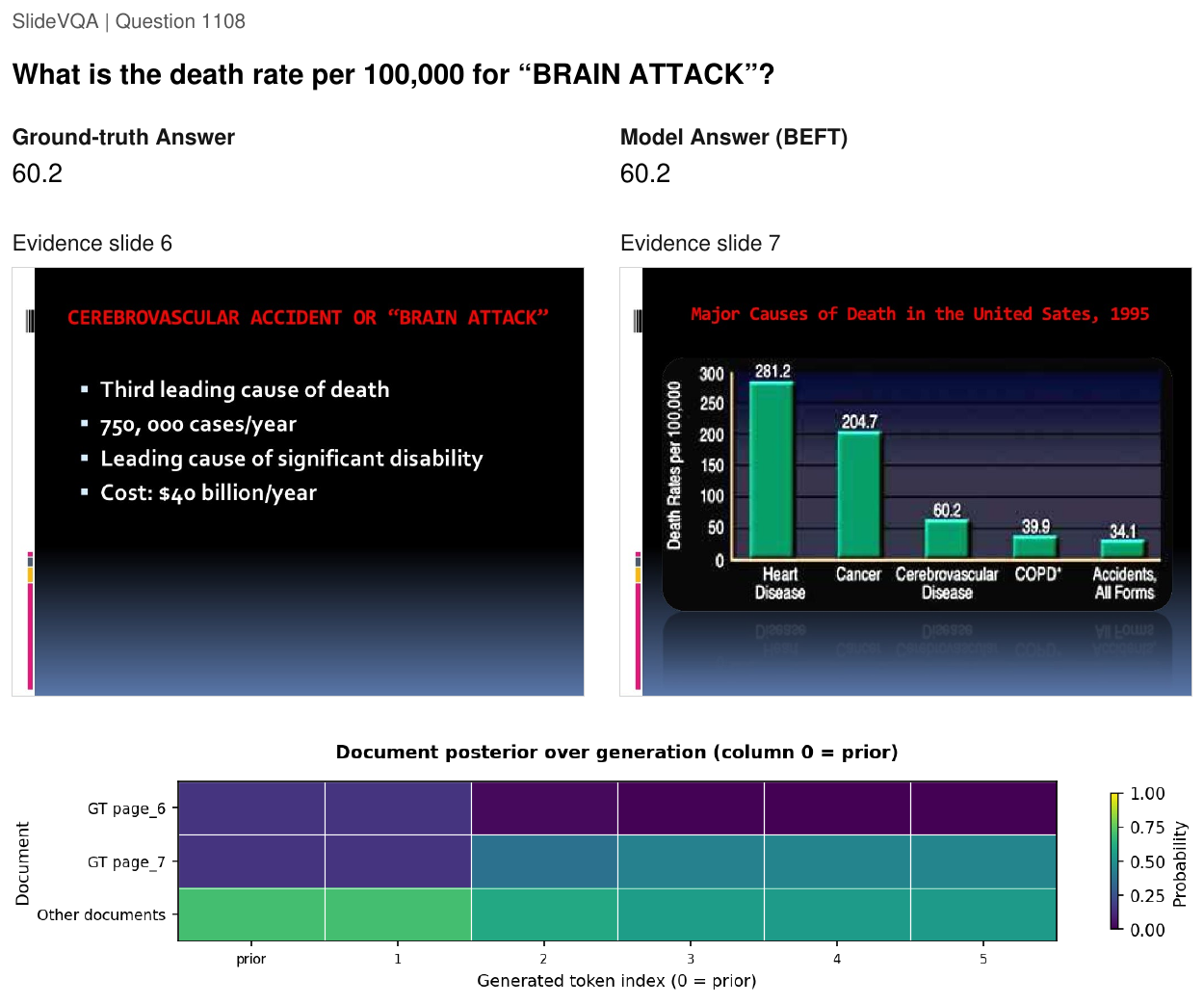}
    \caption{Question ID 1108. Correct answer with multiple ground-truth slides.}
    \label{fig:1108}
\end{figure*}

Figure \ref{fig:1108} shows a correct case with multiple ground-truth slides. The document posterior starts uncertain, but later puts more weight on one of the slides labeled as ground-truth that gives the numerical value needed. The other ground-truth slide provides context for ``Brain Attack''.

\begin{figure*}[!ht]
    \centering
    \includegraphics[width=1.0\linewidth]{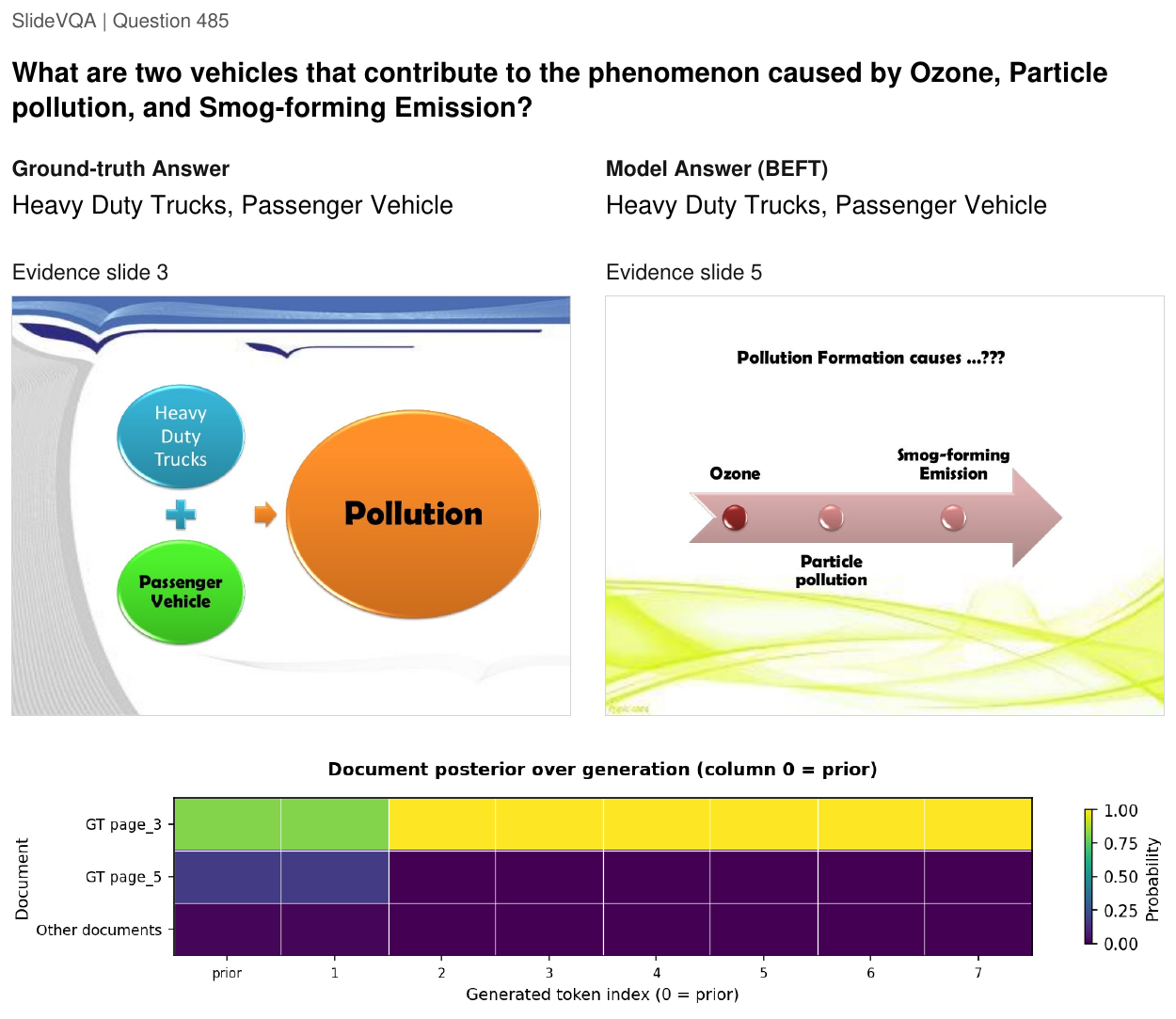}
    \caption{Question ID 485. Correct answer with multiple ground-truth slides.}
    \label{fig:485}
\end{figure*}

Figure \ref{fig:485} shows another correct case with multiple ground-truth slides. The prior put probability mass on both ground-truth document, in which case the model is evaluated as successful on the evidence selection task. Then, as generation proceeds, the posterior focuses on page 3, where the actual answers are located.

\begin{figure*}[!ht]
    \centering
    \includegraphics[width=1.0\linewidth]{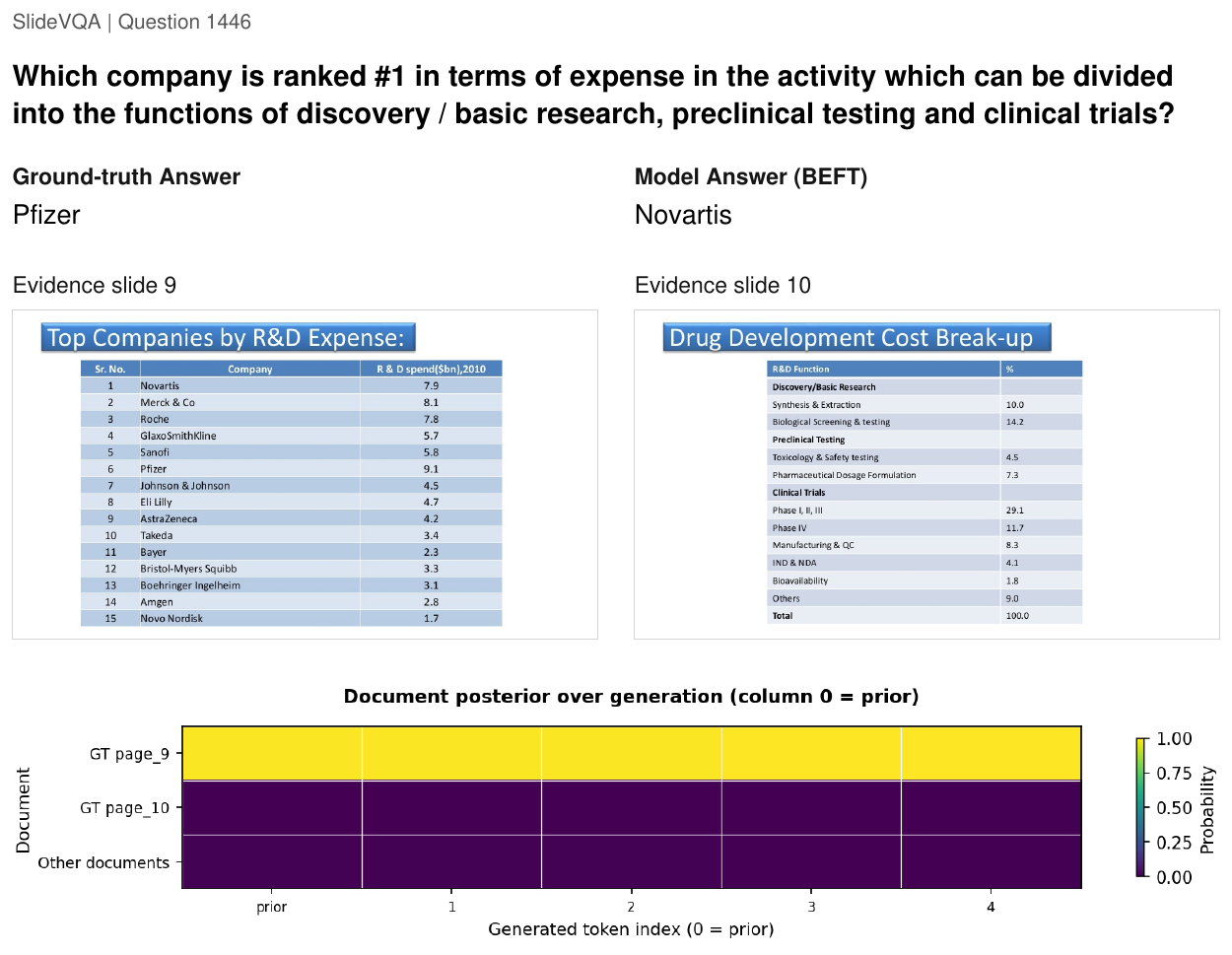}
    \caption{Question ID 1446. Wrong answer under multiple ground-truth slides. This is likely due to model mistakenly treating the left-hand-side index as a rank.}
    \label{fig:1446}
\end{figure*}

Figure \ref{fig:1446} shows an error case under multiple ground-truth slides. We find that the posterior focuses on the correct slide with the relevant information. However, the model likely mistakenly treats the left-hand-side index on page 9 as a rank, where the right-hand-side score on the same page should be used to find the top-ranked company. Again, such detailed error attribution is made possible with the concentrated document posterior.

\end{document}